\documentclass{article}

    \usepackage[preprint]{neurips_2025}

\usepackage[dvipsnames, table, HTML]{xcolor}
\usepackage[colorlinks=true,
            linkcolor=teal,
            citecolor=teal,
            urlcolor=MidnightBlue]{hyperref}
\usepackage[utf8]{inputenc} % allow utf-8 input
\usepackage[T1]{fontenc}    % use 8-bit T1 fonts
\usepackage{hyperref}       % hyperlinks
\usepackage{url}            % simple URL typesetting
\usepackage{booktabs}       % professional-quality tables
\usepackage{amsfonts}       % blackboard math symbols
\usepackage{nicefrac}       % compact symbols for 1/2, etc.
\usepackage{microtype}      % microtypography
\usepackage{tcolorbox}
\usepackage{highlight}
\usepackage{blindtext}
\usepackage{multicol}
\usepackage{multirow}
\usepackage{enumitem}
\usepackage{wrapfig}
\usepackage{amsmath}
\usepackage{subcaption}
\definecolor{forest}{RGB}{34,139,34}     
\usepackage{booktabs}      % \toprule, \midrule, \bottomrule
\usepackage{subcaption}    % subtable environment & \captionof

\usepackage{booktabs}
\usepackage{subcaption}
\newlength\savewidth

\newcommand{\tablestyle}[2]{\setlength{\tabcolsep}{#1}\renewcommand{\arraystretch}{#2}\centering\footnotesize}
\usepackage{booktabs}      % \toprule, \midrule, \bottomrule
\usepackage{multirow}      % \multirow
\usepackage{tocbibind}  % For adding things to TOC
\usepackage{titletoc}   % For partial TOC

\usepackage[font=footnotesize,labelfont=bf,aboveskip=2pt,belowskip=2pt]{caption}
\newcommand{\benchmarkname}{{MathIF}\xspace}
\definecolor{skyblue}{RGB}{0, 102, 204}

\NewDocumentCommand{\yafu}
{ mO{} }{\textcolor{red}{\textsuperscript{\textit{yafu}}\textsf{\textbf{\small[#1]}}}}

\definecolor{IFcol}{HTML}{EDF6F7}   % Sky
\definecolor{Ccol}{HTML}{FFF6EF}    % Sage
\definecolor{colA}{HTML}{EDF6F7}   % Sky
\definecolor{colB}{HTML}{FFF6EF}    % Sage
\definecolor{colC}{HTML}{bee9e8}    % Sage
\definecolor{colD}{HTML}{ffccd5}    % Sage
\definecolor{colE}{HTML}{80ed99}    % Sage

\title{Scaling Reasoning, Losing Control: Evaluating Instruction Following in Large Reasoning Models}

\author{%
  Tingchen Fu$^{1}$, Jiawei Gu$^{2}$, \textbf{Yafu Li}$^{2}$\thanks{Corresponding author: yafuly@gmail.com}\hspace{0.5mm}, Xiaoye Qu$^{2}$, Yu Cheng$^{3}$ \\
  $^1$ Renmin University of China,  $^2$ Shanghai AI Laboratory, \\
  $^3$ The Chinese University of Hong Kong \\
}

\begin{document}

\maketitle

\begin{abstract}

Instruction-following is essential for aligning large language models (LLMs) with user intent. 
While recent reasoning-oriented models exhibit impressive performance on complex mathematical problems, their ability to adhere to natural language instructions remains underexplored. 
In this work, we introduce \textbf{MathIF}, a dedicated benchmark for evaluating instruction-following in mathematical reasoning tasks. 
Our empirical analysis reveals a consistent tension between scaling up reasoning capacity and maintaining controllability, as models that reason more effectively often struggle to comply with user directives.
We find that models tuned on distilled long chains-of-thought or trained with reasoning-oriented reinforcement learning often degrade in instruction adherence, especially when generation length increases. 
Furthermore, we show that even simple interventions can partially recover obedience, though at the cost of reasoning performance. 
These findings highlight a fundamental tension in current LLM training paradigms and motivate the need for more instruction-aware reasoning models. 
We release the code and data at \href{https://github.com/TingchenFu/MathIF}{https://github.com/TingchenFu/MathIF}.
% We conclude with recommendations for evaluating and training instruction-aligned reasoning models.
\end{abstract}

\section{Introduction}

% LRMs strong reasoning performance; smarter -> 
Recent advancements in Large Reasoning Models (LRMs)~\cite{qu2025survey}, such as o3 and o4-mini~\cite{o3}, DeepSeek-R1~\cite{deepseek-r1}, and K1.5~\cite{kimi}, have demonstrated impressive capabilities in mathematical reasoning, including solving olympiad-level problems~\cite{OlympiadBench,math500,aime_1983_2024} and automating formal theorem proving~\cite{deepseek-math-prover}. These breakthroughs have sparked growing interest in scaling chain-of-thought (CoT) reasoning~\cite{cot}, where models produce explicit multi-step explanations to solve complex tasks. Typical approaches include imitation learning, e.g.,  supervised fine-tuning (SFT), and reinforcement learning with verifiable rewards~\cite{rl-vr}, both of which aim to strengthen model intelligence across various tasks and scales.
% instruction following; less discussed; we found weak

Despite these advances, instruction following, i.e., the ability to accurately and reliably comply with user directives, has received comparatively little attention in the context of LRMs. 
Yet this ability is critical for real-world alignment and safety~\cite{surveyllmasajudge}. 
Our empirical evaluations on IFEval~\cite{ifeval} and FollowBench~\cite{followbench} reveal a consistent pattern: although LRMs excel at mathematical reasoning, they often fail to follow even simple instructions. 
This raises an important question: 
% \textbf{\textit{Does improving reasoning capability inherently compromise a model’s ability to follow user instructions?}}
% \textbf{\textit{As reasoning scales, do we gain more intelligent models but lose control over their behavior?}}
\textbf{\textit{As reasoning scales, do models become more intelligent yet less controllable?}}
% As reasoning scales, is control lost? Does improved reasoning capability come at the cost of reduced instruction fidelity?
% Unfortunately, existing instruction-following benchmarks are ill-suited for this question in reasoning-heavy contexts. Most are designed for general-purpose language models and overlook math-specific instruction behaviors. By contrast, LRMs are trained predominantly on math-centric datasets and optimized for problem-solving, creating a domain mismatch that obstructs a clean assessment of instruction adherence.
% This gap highlights the urgent need to evaluate whether increasing intelligence in specialized reasoning models inherently leads to diminishing control over their behavior—an issue at the heart of instruction alignment in the age of advanced LRMs.
Unfortunately, existing instruction-following benchmarks are ill-suited for answering this question. 
Most are designed for general-purpose language models and lack coverage of math-specific reasoning behaviors. 
In contrast, LRMs are typically trained on math-heavy datasets and optimized specifically for problem-solving capacity. 
This gap highlights the urgent need to evaluate whether increasing intelligence in specialized reasoning models inherently leads to diminishing control over their behavior, an issue at the heart of instruction alignment for advanced LRMs.

To address this, we propose \textbf{\benchmarkname}, the first benchmark specifically designed to evaluate the instruction-following capabilities of LRMs in math domains. \benchmarkname introduces 15 Python-verifiable constraints across 4 categories, which are programmatically combined to create 30 double-constraint and 15 triple-constraint prompts. 
These are applied to math problems of varying difficulty, resulting in a total of 420 high-quality evaluation samples.
Using \benchmarkname, we evaluate 23 LRMs across a wide range of scales and architectures. 
Surprisingly, most models fail to reliably follow instructions, and performance does not consistently improve with larger model sizes. Even the best-performing model, Qwen3-14B, achieves only 50.71\% accuracy on strict instruction-following. Furthermore, performance deteriorates with increasing task difficulty and constraint complexity, revealing substantial headroom for improvement.

% With our proposed \benchmarkname, we benchmark $23$ recently released LRMs of varying size and architectures. Surprisingly, our experiments reveal that most LRMs fail to align with user intent effectively, and the instruction-following ability does not necessarily improve when scaling up parameter sizes. Even as the best-performing model, Qwen3-14B achieves only 50.71\% accuracy of strict instruction-following. Moreover, we find that the instruction-following performance can become even worse on difficult math problems and with more constraints, highlighting significant room for improvement. 

% trade-off
Our deeper analysis further uncovers a mutual interference between instruction-following and reasoning capabilities, observed at both training and inference stages.
Common reasoning-oriented training strategies (e.g., SFT and RL) enhance reasoning ability but degrade instruction adherence.
This degradation becomes more pronounced as the CoT length increases, likely because longer reasoning paths widen the contextual gap between the original instruction and the final answer, making it harder for the model to retain and execute directives.
Conversely, enforcing brevity by limiting CoT length improves instruction-following performance, but at the cost of reasoning depth and accuracy.

These observations reveal a consistent pattern: \textbf{\textit{improving reasoning capability often comes at the cost of instruction adherence}}, suggesting an inherent trade-off between the two abilities. 
This trade-off highlights a crucial challenge in LRM development: training for stronger reasoning alone may undermine alignment, and future methods must account for this tension to build models that are both capable and controllable.
% Moreover, through an error analysis, we find that there exists a trade-off between mathematical reasoning and instruction-following, 
% Moreover, further analysis reveals the mutual interference effect between mathematical reasoning and instruction-following capabilities in LRMs at both the training phase and the inference phase. Specifically, we find that common reasoning-oriented training pathways such as SFT (with subsequent RL) and cold-start RL improve reasoning but harm instruction-following ability, especially for SFT on distilled long CoT.  
% putting a limitation on the length of CoT leads to a reasoning performance drop and enhancement in instruction-following, suggesting an intrinsic trade-off between the two abilities. 
% contribution bullet points
\vspace{4pt}
\noindent To summarize, our contributions are three-fold:
\begin{itemize}
\item We introduce \benchmarkname, the first benchmark for systematically measuring instruction-following performance in large reasoning models within math domains.
\item We evaluate 23 recent LRMs and uncover a widespread inability to follow user constraints, especially on harder problems and multi-constraint settings.
\item We identify and empirically validate a \textit{trade-off} between reasoning performance and instruction-following ability, with mutual interference observed during both training and inference.
\end{itemize}

\section{Related Work}

\subsection{Large Reasoning Models (LRMs)}
Recent advances in enhancing the reasoning ability of language models and reimplementing large reasoning models generally fall into two paradigms. 
The first paradigm constructs high-quality long CoT by distilling from more capable LRMs or combining primitive reasoning actions~\cite{2025s1,deepseek-r1}, on which SFT and subsequent RL are conducted. For example, s1~\cite{2025s1} shows that even a small amount of CoT data could significantly promote the reasoning ability, and LIMO~\cite{ye2025limoreasoning} further put forward the less-is-more reasoning hypothesis, stating that sophisticated reasoning capabilities can emerge through minimal demonstration of cognitive processes if relevant domain knowledge is already encoded in the language model during pre-training. Similarly, \cite{yeo2025demystifying} suggests that injecting more cognitive behavior into the long CoT during SFT brings promotion in mathematical benchmarks. 

On the other hand, cold-RL on base language models directly attracts more and more attention in the subfield because of the success of deepseek-R1-zero and the previous findings that the model tends to memorize training data during the SFT process~\cite{chu2025sft}. In contrast with SFT, cold-RL does not rely on long CoT data and provides supervision signals by rewards on the final outcome~\cite{deepseek-r1} or the reasoning process~\cite{liu2025oatzero}. To simplify and accelerate the RL process, various techniques have been proposed, such as dynamic sampling~\cite{yu2025dapo}, process-reward~\cite{prime}, off-policy guidance~\cite{luffy}, and CoT preference optimization~\cite{yang2025thinking}. To compare two different pathways and evaluate the capacity of LRMs~\cite{yeo2025demystifying}, previous studies mostly focus on their complex reasoning performance on olympiad-level math problems, while their instruction-following ability when solving math problems is left unconsidered.

\subsection{Instruction-following benchmarks}
As a crucial factor determining the practicality of a language model for real-world scenarios, the instruction-following ability is a core metric for language model evaluation, with numerous protocols and benchmarks being developed~\cite{dubois2023alpacafarm,chiang2023vicuna}. Earlier benchmarks primarily focused on the completeness of user queries and depended on proprietary language models~\cite{dubois2023alpacafarm,chiang2023vicuna} to measure the response quality by its win-rate over the baseline method, which is an oversimplification of real user queries. For a more comprehensive evaluation, sophisticated benchmarks have been developed to test the ability of a language model in following format constraints~\cite{ifeval,xia2024fofo,tang2024struc}, multi-turn instruction~\cite{he2024multi, li2025structflowbench, han2025can, sirdeshmukh2025multichallenge}, long-context instruction~\cite{wu2024lifbench}, multi-lingual instruction~\cite{he2024multi,li2025xifbench}, compositional instruction~\cite{zhang2025iheval,hayati2025chain,han2025can} and refutation instructions~\cite{yan-etal-2024-refutebench,yan2025refutebench20agentic}. However, most instruction-following benchmarks concentrate on the general domain and relatively straightforward queries. The domain shift and the lack of long CoT become a deterrent for using the benchmark on LRMs.

% [format] Instruction-following evaluation for large language models
% [composition] Chain-of-instructions: Compositional instruction tuning on large language models.
% [format] Fofo: A benchmark to evaluate llms’ format-following capability
% [format] Struc-bench: Are large language models good at generating complex structured tabular data?
% [composition] Can large language models understand real-world complex instructions?
% [multi-turn] Multi-if: Benchmarking llms on multi-turn and multilingual instructions following
% [long context ]LIFBench: Evaluating the Instruction Following Performance and Stability of Large Language Models in Long-Context Scenarios
% [multi-lingual]XIFBench: Evaluating Large Language Models on Multilingual Instruction Following
% [mutli-turn]StructFlowBench: A Structured Flow Benchmark for Multi-turn Instruction Following
% [composition] IHEval: Evaluating language models on following the instruction hierarchy
% [multi-turn]Can Language Models Follow Multiple Turns of Entangled Instructions?
% [multi-turn ]MultiChallenge: A Realistic Multi-Turn Conversation Evaluation Benchmark Challenging to Frontier LLMs 
%
\section{\benchmarkname}

\begin{wrapfigure}{r}{0.5\textwidth}
  \centering
  \vspace{-10pt}
  \includegraphics[width=\linewidth]{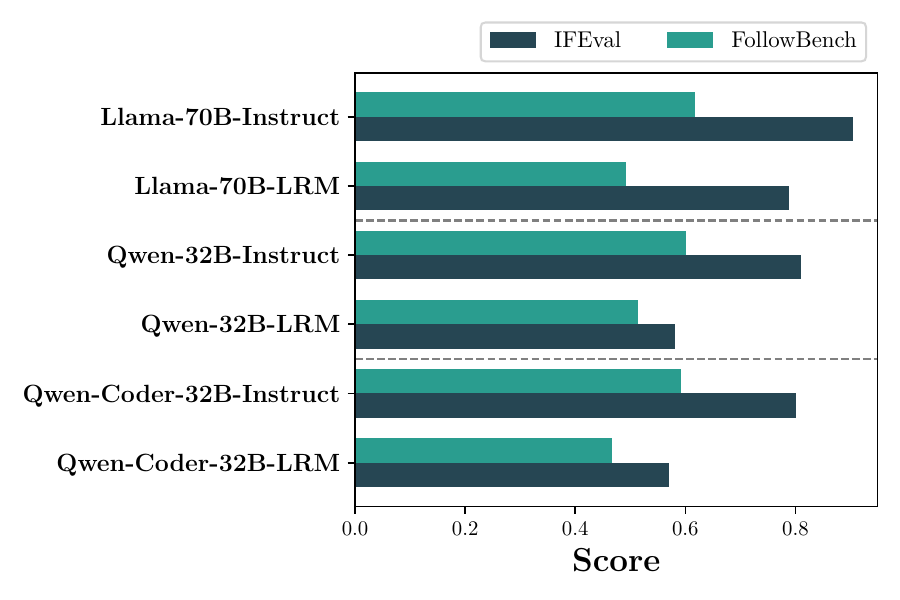}
  % \vspace{-10pt}
  \caption{Performance of Instruct LLMs and LRMs on IFEval~\cite{ifeval} and FollowBench~\cite{followbench}.   
  % , different from the direct query-response mapping of traditional supervised fine-tuning.
  }
  \label{fig:ifeval}
  \vspace{-5pt}
\end{wrapfigure}

% \yafu{Think we can start from general IF scores to give an intuition of how LRMs behave and why we compose MathIF}

\paragraph{Overview}
As shown in Figure~\ref{fig:ifeval}, the same base model exhibits a clear performance drop in both IFEval~\cite{ifeval} and FollowBench~\cite{followbench} after transitioning from the Instruct variant to the LRM variant. 
The ``Instruct'' and ``LRM'' are abbreviations for Instruction LLMs and their corresponding large reasoning models, with details presented in the Appendix. 
This observation suggests that reasoning-oriented training, while beneficial for problem-solving, may compromise instruction-following ability. 
However, since both benchmarks are designed for general-purpose tasks rather than mathematical reasoning, where LRMs are specifically optimized, it remains difficult to isolate instruction-following performance from confounding factors such as domain mismatch.

In this study, we propose \benchmarkname, the first benchmark for evaluating the instruction-following ability of LRMs. Specifically, in this benchmark, we consider four major types of Python-verifiable constraints and construct compositional instructions by combining two or three constraints together. The constraints are then incorporated into the math problems, which are collected from diverse sources and vary in levels of difficulty. Two metrics are proposed for fine-grained analysis of LRM's instruction-following ability.   

\begin{table}[t]
  \centering
  \small
  \setlength{\tabcolsep}{5pt}
  \renewcommand{\arraystretch}{0.7}
  \caption{Single constraints and sample dual-/triple-constraint compositions across four categories.}
  \label{tab:constraint_examples}
  \vspace{1pt}
  \resizebox{\textwidth}{!}{%
    \begin{tabular}{%
      >{\bfseries}l  % 第一列（Category）整列加粗
      l               % 第二列（Sub-Category）正常体，仅表头加粗
      >{}p{0.8\linewidth} % 第三列小号打字机体
    }
      \toprule
      Category    & {\bfseries Sub-Category}
                  & {\normalfont\bfseries Example} \\
      \midrule
      \multirow[t]{1}{*}{\raisebox{0.1ex}[0pt][0pt]{Length}}
        & Length        
        & Answer with less than 500 words. \\
      \midrule
      \multirow[t]{2}{*}{Lexical}
        & Language      
        & Your answer should be in \texttt{Chinese language}, no other language is allowed. \\
        & Keyword       
        & Include keywords \texttt{"condition"} in your response. \\
      \midrule
      \multirow[t]{3}{*}{Format}
        & Punctuation   
        & In your entire response, refrain from the use of any commas. \\
        & Case          
        & Your entire response should be in English, and in all lowercase letters. No capital letters are allowed. \\
        & Highlight     
        & Your answer must contain exactly 3 bullet points. Use the markdown bullet points such as:
                          * This is point 1.
                          * This is point 2. \\
      \midrule
      \multirow[t]{3}{*}{Affix}
        & Prefix        
        & First repeat the request word for word without change, then give your answer. \\
        & Suffix        
        & Finish your response with this exact phrase \texttt{"Any other questions?"}. No other words should follow this phrase. \\
        & Both          
        & Wrap your entire response with double quotation marks. \\
      \bottomrule
    \end{tabular}%
  } % end resizebox
\end{table}

\paragraph{Constraint Type} 
Inspired by IFEval~\cite{ifeval} and ComplexBench~\cite{complexbench}, we mainly consider $15$ constraints in four types when establishing our benchmark: (1) \textbf{length constraint} puts a limitation on the length of the response since an overlong response might lead to extra latency at inference; (2) \textbf{lexical constraint} requires the model to give the response in a specified language or contains pre-defined key words and phrases in response; (3) \textbf{format constraint}, being probably the most common type of constraint from real users of LRMs, it includes various forms of restriction including the number of sections, the number of highlighted bullet, usage of punctuation, case sensitivity and many other forms of instruction; and finally (4) \textbf{affix constraint}, refers to the prefix constraint or suffix constraint or a combination of both that requires the model to begin/end the response with specific symbols or sentences. To mitigate reliance on proprietary language model~\cite{hurst2024gpt}, we ensure that the compliance of constraints in our benchmark can be verified simply by a Python programme. A more detailed categorization for the type and subtype of constraints involved in our benchmark is listed in Table~\ref{tab:constraint_examples} together with an illustrating example. The entire list of constraints used in our benchmark can be found in the Appendix.

\paragraph{Compositional Constraint} % or compounded constraint?
Queries with only a single constraint can hardly reflect the complex scenarios encountered by a downstream application of LRM, as the real user queries to LRMs typically contain more than one restrictive condition. Therefore, following previous work in complex instruction-following~\cite{complexbench}, we construct compositional constraints by combining two or three individual constraints. Specifically, given the set of individual constraints denoted as $\mathcal{C}$, we enumerate all the elements in the Cartesian product $\mathcal{C}^2 = \{(c_1,c_2) \mid c_1,c_2 \in \mathcal{C}\}$ and $\mathcal{C}^3 = \{(c_1,c_2,c_3) \mid c_1,c_2,c_3 \in \mathcal{C}\}$, from which we randomly sample several combinations after manually filtering out the ones in which the constraints are incompatible with each other and fall into the same subtype of constraint. Through this procedure, we harvest $30$ dual-constraints and $15$ triple-constraints. The detailed list of dual-constraints and triple-constraints is presented in Table~\ref{tab:constraint_examples}.

\paragraph{Math Problem Collection} With the constructed individual constraints and compositional constants, the next step is to incorporate these constraints into math problems to constitute a query. Different from recent benchmarks that mostly center around math problems from undergraduate level or even higher level~\cite{aime_1983_2024}, \benchmarkname contains math problems of varying levels of difficulty, ranging from math word problems in primary school and math problems in high school to the latest math problems in world-level competition. Specifically, we randomly sample $90$ problems from GSM8K~\cite{gsm8k}, MATH-500~\cite{math500}, Minerva~\cite{de2013minerva}, Olympiad~\cite{OlympiadBench} respectively. For AIME2024\&2025~\cite{aime_1983_2024}, we use all the $60$ problems it contained. 
For each data source, we apply a single constraint, dual constraints, and triple constraints, resulting in three subsets of equivalent size. 
% \yafu{What does one-third mean here? Is each math problem assigned with three IF cases for one, two and three constraints respectively?}
For a sanity check, we manually review the curated samples and double-check whether the added constraints are contradictory to the math problem itself. 
The statistics for the established dataset are shown in Table~\ref{tab:dataset_stats}. 

% OMNI-MATH: A UNIVERSAL OLYMPIAD LEVEL MATHEMATIC BENCHMARK FOR LARGE LANGUAGE MODELS
% FRONTIERMATH: A BENCHMARK FOR EVALUATING ADVANCED MATHEMATICAL REASONING IN AI

% \begin{table}[]
%     \centering
%     \resizebox{1.0\linewidth}{!}{
%     \input{table/statistics}
%     }
%     \caption{The statistics of our proposed \benchmarkname grouped by the source of math problems and the number of constraints.}
%     \label{tab:statistics}
% \end{table}

% 导言区（确保已加载）
% \usepackage{booktabs,graphicx,subcaption,multirow}
% \newcommand{\tablestyle}[2]{\setlength{\tabcolsep}{#1}%
%   \renewcommand{\arraystretch}{#2}%
%   \centering\footnotesize}

% 前导区请确保已加载 booktabs, graphicx, multirow, subcaption 等宏包，
% 并定义了 \tablestyle 宏：
% \newcommand{\tablestyle}[2]{\setlength{\tabcolsep}{#1}%
%   \renewcommand{\arraystretch}{#2}%
%   \centering\footnotesize}

\begin{table}[t]
  \centering
  \footnotesize
  \tablestyle{6pt}{0.8}  % 列距 2pt，行距 0.8，且切换到 
  \caption{Dataset statistics grouped by source and by constraint.}
  \label{tab:dataset_stats}
  \vspace{3pt}

  \resizebox{\textwidth}{!}{%
    \begin{tabular}{lccccccccc}
      \toprule
             & \multicolumn{5}{c}{\bfseries Group by source}
             & \multicolumn{3}{c}{\bfseries Group by constraint}
             & \multirow{2}{*}{\bfseries Total} \\
      \cmidrule(lr){2-6}\cmidrule(lr){7-9}
             & \bfseries GSM8K & \bfseries MATH500 & \bfseries Minerva
             & \bfseries Olympiad & \bfseries AIME
             & \bfseries Single & \bfseries Double & \bfseries Triple
             & \\
      \midrule
      \bfseries \# samples 
           &  90   &   90   &   90   &   90     &   60
           &  140  &  140   &  140   &  420    \\
      \bfseries Avg.\ Len 
           & 86.73 & 57.24  & 88.09  & 80.42    & 87.25
           & 64.89 & 83.84  & 89.54  & 79.43   \\
      \bottomrule
    \end{tabular}%
  } % end resizebox
\end{table}

\paragraph{Evaluation Metric} To systematically measure whether one or more constraints in the query are satisfied by the LRM while solving the math problems, we follow previous works~\cite{ifeval,followbench} and use two metrics of different granularity. Specifically, we employ \textbf{hard accuracy (HAcc)} and \textbf{soft accuracy (SAcc)} to measure whether the model response follows the constraints at the query level and constraint level, respectively. Formally, suppose a query has $n$ constraints $\mathcal{C}_1, \mathcal{C}_2, \mathcal{C}_3, \ldots, \mathcal{C}_n$ and we use $\mathbb{I}(\mathcal{C}_i)$ to denote whether the $i$-th constraint is satisfied or not, with $\mathbb{I}(\mathcal{C}_i)=1$ for satisfied constraint and $\mathbb{I}(\mathcal{C}_i)=0$ for unsatisfied constraint. The hard accuracy (HAcc) and soft accuracy (SAcc) for a query is defined as:
\begin{equation}
    {\rm{HAcc}} = \prod_{i=1}^n \mathbb{I}(\mathcal{C}_i), \quad {\rm{SAcc}} = \frac{1}{n}\sum_{i=1}^n \mathbb{I}(\mathcal{C}_i)
\end{equation}
% And the soft accuracy (SAcc) is defined as:
% % \yafu{Should be SAcc in the following eq.?}
% \begin{equation}
    
% \end{equation}

Notably, for queries with only a single constraint, the two metrics are identical in number.  The overall hard accuracy and soft accuracy on the benchmark are averaged among all the queries in the dataset. Apart from instruction-following ability, we also measure the correctness of the math problem solution on our proposed \benchmarkname, defined as whether the final answer exactly matches the ground-truth, regardless of constraint satisfaction.
By default, correctness refers to performance with constraints in the prompts unless specified (e.g., Table~\ref{tab:main}).

\section{Experiment}
\label{sec:exp}

% \paragraph{Model Configuration.}

To benchmark the instruction-following ability of LRMs, we evaluate a diverse set of models across three parameter scales. All models are decoded using nucleus sampling ($T{=}1.0$, $p{=}0.95$) with a maximum generation length of 16,384 tokens, powered by the \texttt{vLLM}~\cite{vllm} engine for efficient inference.

% \yafu{Use official names, capitalize where necessary, and align with the table.}

\begin{itemize}[wide=0.\parindent,noitemsep,topsep=0.em]
    \item \textbf{Small-scale models ($\leq$ 4B parameters):}  
    Qwen3-0.6B~\cite{qwen3}, Qwen2.5-1.5B-SimpleRL-Zoo~\cite{simplerl}, Qwen2.5-Math-1.5B-Instruct~\cite{qwen-math}, DeepSeek-R1-Distill-Qwen-1.5B~\cite{deepseek-r1}, DeepScaler-1.5B-Preview~\cite{deepscaler2025}, L1-Qwen-1.5B-Max~\cite{l1}, L1-Qwen-1.5B-Exact~\cite{l1}, Qwen3-1.7B~\cite{qwen3}, Qwen3-4B~\cite{qwen3}.
    
    \item \textbf{Medium-scale models (7B$\sim$14B parameters):}  
    Qwen2.5-Math-7B-Instruct~\cite{qwen-math}, DeepSeek-R1-Distill-Qwen-7B~\cite{deepseek-r1}, Open-Reasoner-Zero-7B~\cite{hu2025open}, DeepSeek-R1-Distill-Llama-8B~\cite{deepseek-r1}, Qwen3-8B~\cite{qwen3}, DeepSeek-R1-Distill-Qwen-14B~\cite{deepseek-r1}, Qwen3-14B~\cite{qwen3}.
    
    \item \textbf{Large-scale models ($\geq$ 32B parameters):}  
    s1-32B~\cite{2025s1}, OlympicCoder-32B~\cite{openr1}, DeepSeek-R1-Distill-Qwen-32B~\cite{deepseek-r1}, QwQ-32B~\cite{qwq32b}, Open-Reasoner-Zero-32B~\cite{reasoner-zero}, Qwen3-32B~\cite{qwen3}, DeepSeek-R1-Distill-Llama-70B~\cite{deepseek-r1}.
\end{itemize}

\subsection{Experimental Results}
\label{sec:exp_res}

\begin{table}[t]
  \centering
  % --- 保留原来的 caption 和 label ---
  \small
  % 将字体与行列间距改为第一张表格的设置：
  \tablestyle{6pt}{1.1}
  \caption{
    Experimental results of LRMs on \benchmarkname. We report hard accuracy (HAcc) and soft accuracy (SAcc) for instruction-following, alongside math-solving correctness \textit{with} and \textit{without} constraints (w/o const. / w/ const.). The last column shows the relative change in correctness when constraints are included. Models are sorted in descending order of instruction-following performance. $\dagger$ indicates models trained by supervised fine-tuning only (no reasoning-oriented RL). 
    \textbf{Bold} and \underline{underlined} values denote the \textit{top}-2 and \textit{bottom}-2 entries in each column, respectively.
  }
  \label{tab:main}
  \vspace{3pt}

  % 保持原来的 0.9\textwidth 缩放
  \resizebox{0.9\textwidth}{!}{%
    \begin{tabular}{
      l
      >{\columncolor{IFcol}}c
      >{\columncolor{IFcol}}c
      >{\columncolor{Ccol}}c
      >{\columncolor{Ccol}}c
      >{\columncolor{Ccol}}c
    }
      \toprule
      \multirow{2}{*}{\textbf{Model}}
       & \multicolumn{2}{>{\columncolor{IFcol}}c}{\textbf{Instruction Following}}
       & \multicolumn{3}{>{\columncolor{Ccol}}c}{\textbf{Correctness}} \\
                & \textbf{HAcc}                & \textbf{SAcc}                & \textbf{w/o const.}        & \textbf{w/ const.}        & \textbf{Diff.(\%)}   \\
      \midrule
      \multicolumn{6}{c}{\textit{Models with no more than 4B parameters}} \\
      \midrule
      Qwen3-4B                      
        & \textbf{44.05} & \textbf{61.43} & \textbf{68.10} & \textbf{58.57} & \textbf{-13.99}  \\
      Qwen3-1.7B                    
        & \textbf{30.24} & 50.24          & \textbf{62.38} & \textbf{51.19} & -17.94  \\
      Qwen3-0.6B                    
        & 27.86          & \textbf{50.44} & \underline{40.95} & 32.14       & -21.51  \\
      L1-Qwen-1.5B-Exact            
        & 19.76          & 39.60          & 53.81          & 42.86           & -20.35  \\
      L1-Qwen-1.5B-Max              
        & 19.76          & 39.40          & 55.48          & 45.71           & -17.61  \\
      DeepSeek-R1-Distill-Qwen-1.5B$\dagger$
        & 17.14          & 36.62          & 52.86          & \underline{31.67} & \underline{-40.09}  \\
      DeepScaler-1.5B-Preview       
        & 14.52          & 34.52          & 58.10          & 36.19           & \underline{-37.71}  \\
      Qwen2.5-1.5B-SimpleRL-Zoo     
        & \underline{9.05}  & \underline{24.33} & \underline{27.14} & \underline{22.38} & -17.54  \\
      Qwen2.5-Math-1.5B-Instruct    
        & \underline{7.62}  & \underline{21.39} & 44.05          & 44.29           & \textbf{+0.54}   \\
      \midrule
      \multicolumn{6}{c}{\textit{Models with approximately 7B–14B parameters}} \\
      \midrule
      Qwen3-14B                     
        & \textbf{50.71}          & \textbf{67.06}          & 71.43          & \textbf{64.29}           & -10.00  \\
      DeepSeek-R1-Distill-Qwen-14B$\dagger$
        & \textbf{39.28}          & \textbf{60.55}          & 67.14          & 50.95           & -24.11  \\
      Qwen3-8B                      
        & 37.86          & 57.34          & \textbf{69.52}          & \textbf{66.43}           & \textbf{-4.44}   \\
      DeepSeek-R1-Distill-Qwen-7B$\dagger$
        & 26.43          & 44.96          & 65.24          & 48.57           & \underline{-25.55}  \\
      DeepSeek-R1-Distill-Llama-8B$\dagger$
        & 22.14          & 44.04          & 59.76          & \underline{36.43}           & \underline{-39.04}  \\
      Open-Reasoner-Zero-7B         
        & \underline{13.57}          & \underline{32.26}          & \underline{52.86}          & 51.90           & \textbf{-1.82}   \\
      Qwen2.5-Math-7B-Instruct      
        & \underline{9.05}           & \underline{25.60}          & \underline{46.90}          & \underline{37.14}           & -20.81  \\
      \midrule
      \multicolumn{6}{c}{\textit{Models with 32B or more parameters}} \\
      \midrule
      Qwen3-32B                     
        & \textbf{43.81}          & \textbf{62.82}          & \textbf{72.62}          & \textbf{70.00}           & -3.61 \\
      DeepSeek-R1-Distill-Qwen-32B$\dagger$
        & \textbf{42.62}          & 60.91          & \textbf{71.43}          & 57.62           & \underline{-19.33}  \\
      DeepSeek-R1-Distill-Llama-70B$\dagger$
        & 41.43          & \textbf{61.07}          & 71.19          & \underline{54.05}           & \underline{-24.08}  \\
      QwQ-32B                       
        & 40.24          & 59.99          & 70.95          & \textbf{68.81}           & \textbf{-3.02} \\
      OlympicCoder-32B              
        & 35.95          & 57.97          & \underline{59.29}          & \underline{54.52}           & -8.05 \\
      s1-32B$\dagger$               
        & \underline{20.95}          & \underline{41.78}          & \underline{62.86}          & 60.95           & -3.04 \\
      Open-Reasoner-Zero-32B        
        & \underline{15.47}          & \underline{35.52}          & 65.48          & 67.62           & \textbf{+3.27} \\
      \bottomrule
    \end{tabular}%
  } % end resizebox
\end{table}

\vspace{-10pt}

The experimental results, as summarized in Table~\ref{tab:main}, reveal several key factors that influence the instruction-following performance of LRMs, including model scale, architecture design, and reasoning format.

% \yafu{first discuss all models achieve relatively low IF performance?}

\noindent \textbf{All LRMs fail to obey most user instructions.}
All LRMs evaluated on \benchmarkname{} exhibit poor instruction-following performance. Even the best-performing model, Qwen3-14B, achieves only 50.71\% hard accuracy, barely surpassing the halfway mark. 
The majority of models, including large-scale variants such as DeepSeek-R1-Distill-Llama-70B and Open-Reasoner-Zero-32B, fail to meet even minimal expectations for faithfully executing user-specified constraints. 
% This result underscores a critical limitation in current reasoning-focused models: improving task performance alone does not guarantee alignment with user intent. 

\noindent \textbf{The Qwen3 series consistently demonstrates strong instruction-following capabilities across model scales.}
Qwen3-14B achieves the highest hard and soft accuracy among all LRMs. Qwen3-4B leads in the sub-4B category, while Qwen3-32B outperforms all models with 32B or more parameters. Overall, the Qwen3 series shows remarkable performance across sizes.

\noindent \textbf{Model scale alone does not determine instruction-following performance.}
While larger models often perform better within the same series (e.g., Qwen2.5-Math and Open-Reasoner-Zero), scaling up does not guarantee improvement across different architectures. For instance, DeepSeek-R1-Distill-Llama-70B underperforms Qwen3-4B despite being more than $15\times$ larger. Notably, Qwen3-8B and Qwen3-32B deviate from the within-series scaling trend, highlighting that instruction-following ability depends on both model size and design.

\noindent \textbf{Explicit reasoning separation may enhance instruction-following.}
Most LRMs in our study adopt special tokens (e.g., \texttt{<think>} and \texttt{</think>}) to isolate reasoning from the final answer, with only four exceptions: Qwen2.5-Math-{1.5B, 7B}-Instruct and Qwen2.5-{1.5B, 7B}-SimpleRL-Zoo. These models generally underperform their peers of similar size, suggesting that the lack of explicit separation may blur the boundary between reasoning and answer, leading to poorer adherence to constraints.

\noindent \textbf{There exists a trade-off between instruction-following and mathematical reasoning.}
As shown in the ``Diff'' column of Table~\ref{tab:main}, most models experience a drop in problem-solving correctness when additional constraints are introduced, with margins ranging from $0.96$ to $23.33$. This suggests that stronger adherence to external constraints may compromise core mathematical reasoning. The only exceptions are Qwen2.5-Math-1.5B-Instruct and Open-Reasoner-Zero-32B, which maintain or slightly improve their performance under constrained conditions.

\begin{wrapfigure}{r}{0.5\textwidth}
  \centering
  \vspace{-35pt}
  \includegraphics[width=\linewidth]{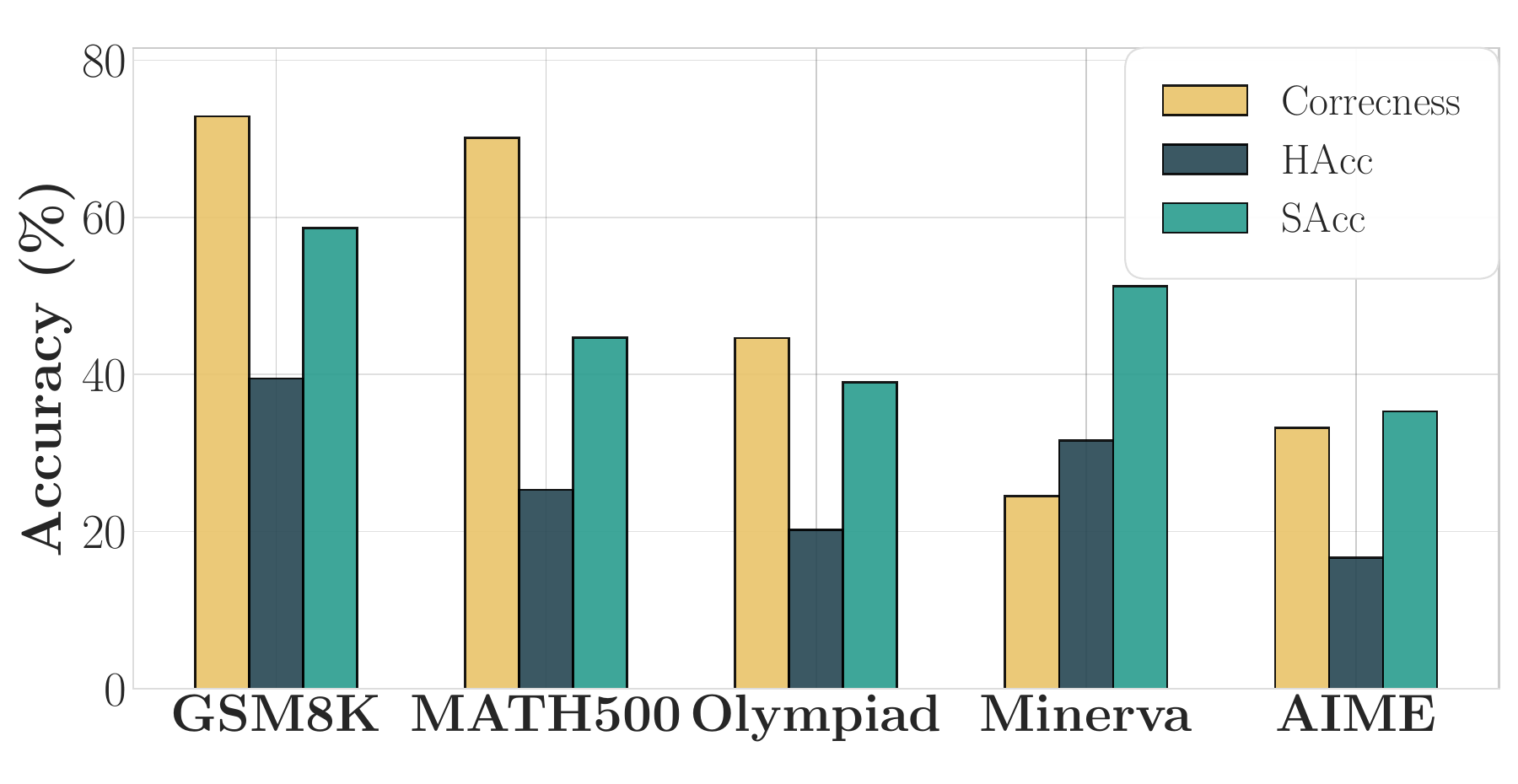}
  % \vspace{-10pt}
  \caption{The accuracy on each subset \benchmarkname averaged over 23 LRMs.   
  % , different from the direct query-response mapping of traditional supervised fine-tuning.
  }
  \label{fig:overall_source}
  % \vspace{-5pt}
\end{wrapfigure}

\subsection{A Closer Look at the Results}
\label{sec:closer}

We first scrutinize the model performance on each subset and visualize the average accuracy of $23$ LRMs in Figure~\ref{fig:overall_source}. From the figures, we can observe a performance difference among different subsets for both hard accuracy and soft accuracy. And the performance on GSM8K, the easiest benchmark among the five, is obviously higher than the other four subsets, especially the challenging AIME subset. Therefore, we can conclude that whether an LRM follows the constraints is correlated with the difficulty level of the math problem and constraints on easier math problems are more likely to be followed. More experimental results could be found in Appendix.

\begin{wrapfigure}{r}{0.4\textwidth}
  \centering
  \vspace{-10pt}
  \includegraphics[width=\linewidth]{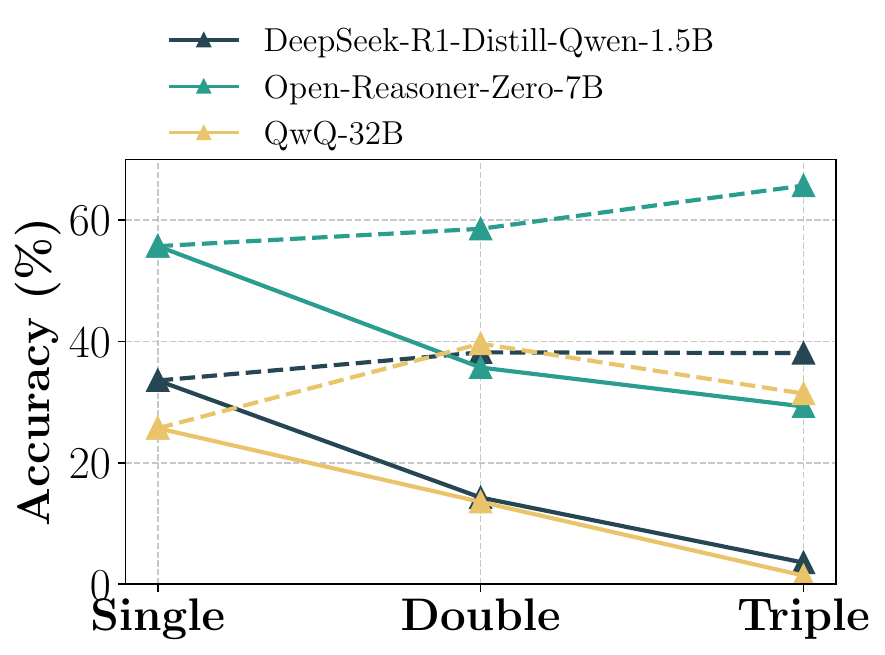}
  % \vspace{-10pt}
  \caption{The HAcc (solid line) and SAcc (dashed line) on the single/double/triple-constraint subset.  
  % , different from the direct query-response mapping of traditional supervised fine-tuning.
  }
  \label{fig:constraint_number}
  \vspace{-10pt}
\end{wrapfigure}

% \begin{minipage}{1.025\textwidth}
% \centering
% \begin{minipage}[h]{0.32\textwidth}
%  \includegraphics[width=1.0\linewidth]{figure/deepseek-r1-distill-qwen-1.5b_constraint_number.pdf}
% % \makeatletter\def\@captype{figure}\makeatother\caption{The accuracy of DeepSeek-R1-Distill-Qwen-1.5B on each subset of \benchmarkname. }
%  \label{fig:fig1}
% \end{minipage}
% \hspace{0.03mm}
% \begin{minipage}[h]{0.32\textwidth}
%  \includegraphics[width=1.0\linewidth]{figure/open-reasoner-zero-7b_constraint_number.pdf}
% % \makeatletter\def\@captype{figure}\makeatother\caption{The accuracy of DeepSeek-R1-Distill-Qwen-1.5B on each subset of \benchmarkname. }
%  \label{fig:fig1}
% \end{minipage}
% \hspace{0.03mm}
% \begin{minipage}[h]{0.32\textwidth}
%  \includegraphics[width=1.0\linewidth]{figure/qwq-32b_constraint_number.pdf}
% % \makeatletter\def\@captype{figure}\makeatother\caption{The accuracy of DeepSeek-R1-Distill-Qwen-1.5B on each subset of \benchmarkname. }
%  \label{fig:fig1}
% \end{minipage}
% \makeatletter\def\@captype{figure}\makeatother\caption{The accuracy of DeepSeek-R1-Distill-Qwen-1.5B (left), Open-Reasoner-Zero-7B (middle) and QwQ-32B (right) on subsets of \benchmarkname with different number of constraints.
% \yafu{Think line plot would fit better here: solid and dashed lines to distinguish H/SAcc while line colors denoting models.}
% }
% \label{fig:constraint_number}
% \end{minipage}

Next, we investigate the impact of the constraint number and plot the instruction-following accuracy of three LRMs in Figure~\ref{fig:constraint_number}. Unsurprisingly, more constraints are associated with the drop in hard accuracy. 
To be more specific, for all the $23$ models involved in our experiment, we can observe an obvious deterioration in hard accuracy when increasing the number of constraints to $2$ (double-constraint) and $3$ (triple-constraint). However, it is worth noting that the same pattern is not applicable to the soft accuracy, as the soft accuracy remains unchanged or even exhibits growth when more constraints are applied. It seems that the model's ability to follow every individual constraint can be enhanced by the existence of more constraints. 
%Meanwhile, for the double-constraint (triple-constraint) setting, the hard accuracy is not equal to the square (cube) of the soft accuracy, indicating the intervention between multiple constraints. \yafu{what does this statement mean?}

\begin{minipage}{1.025\textwidth}
\centering
\begin{minipage}[t]{0.24\textwidth}
 \includegraphics[width=1.0\linewidth]{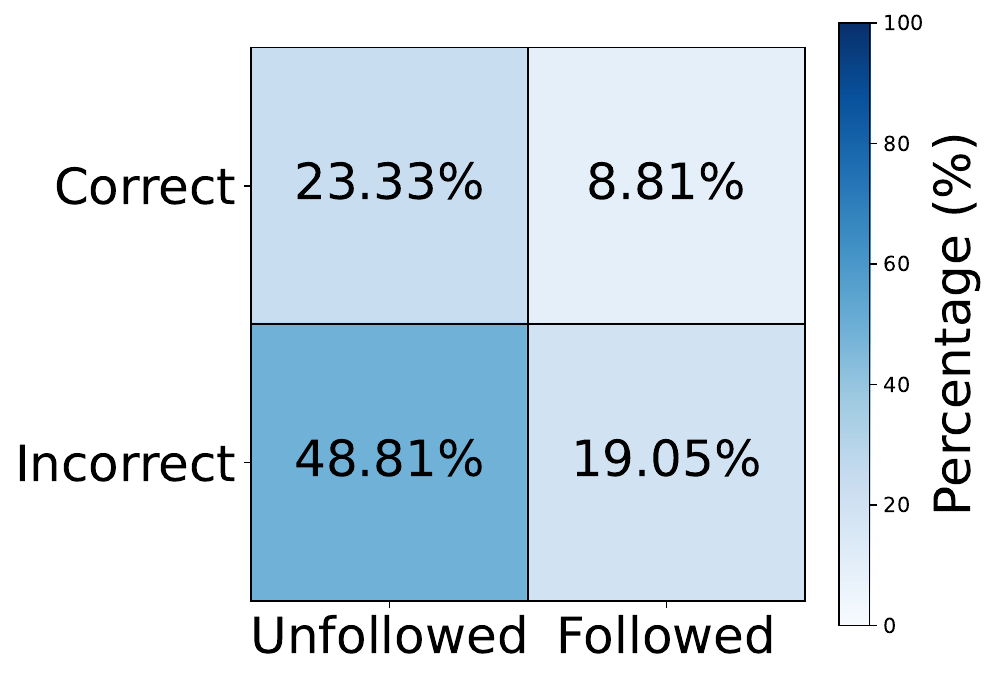}
% \makeatletter\def\@captype{figure}\makeatother\caption{The accuracy of DeepSeek-R1-Distill-Qwen-1.5B on each subset of \benchmarkname. }
 \label{fig:fig1}
\end{minipage}
\begin{minipage}[t]{0.24\textwidth}
 \includegraphics[width=1.0\linewidth]{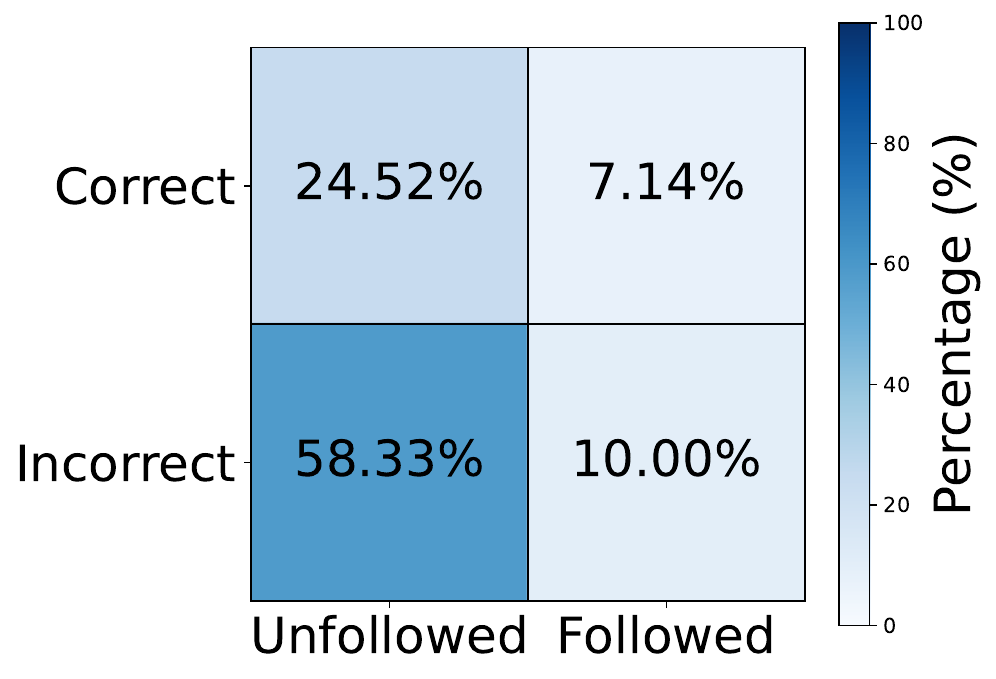}
% \makeatletter\def\@captype{figure}\makeatother\caption{The accuracy of DeepSeek-R1-Distill-Qwen-1.5B on each subset of \benchmarkname. }
 \label{fig:fig1}
\end{minipage}
\hspace{0.03mm}
\begin{minipage}[t]{0.24\textwidth}
 \includegraphics[width=1.0\linewidth]{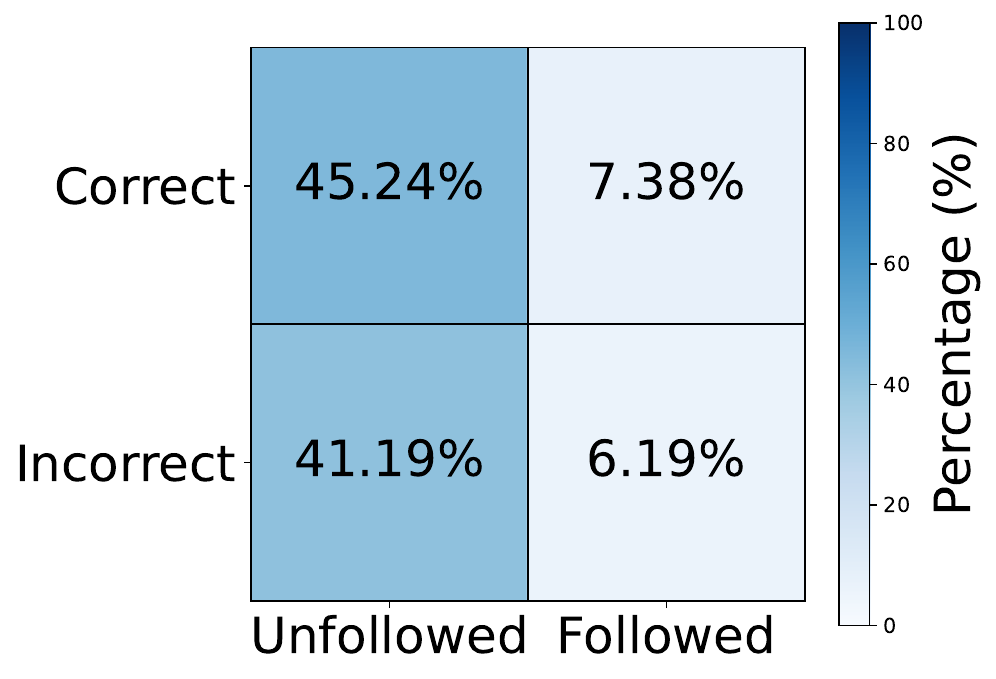}
% \makeatletter\def\@captype{figure}\makeatother\caption{The accuracy of DeepSeek-R1-Distill-Qwen-1.5B on each subset of \benchmarkname. }
 \label{fig:fig1}
\end{minipage}
\begin{minipage}[t]{0.24\textwidth}
 \includegraphics[width=1.0\linewidth]{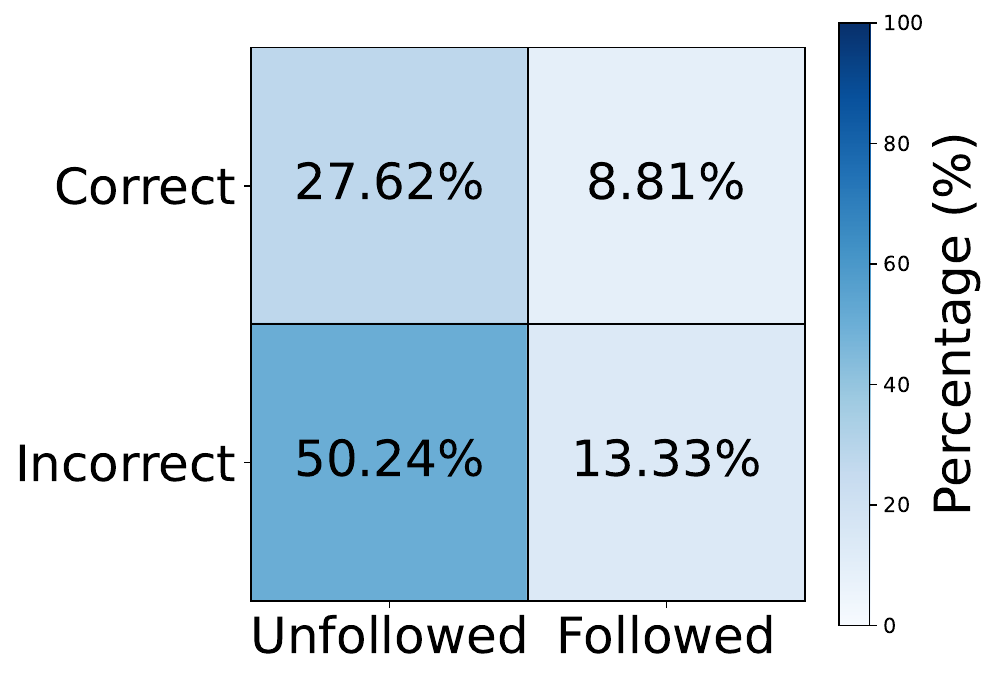}
% \makeatletter\def\@captype{figure}\makeatother\caption{The accuracy of DeepSeek-R1-Distill-Qwen-1.5B on each subset of \benchmarkname. }
 \label{fig:fig1}
\end{minipage}
\hspace{0.03mm}
\makeatletter\def\@captype{figure}\makeatother\caption{
Error set analysis for Qwen3-0.6B, DeepSeek-R1-Distill-Qwen-1.5B, Open-Reasoner-Zero-7B, and DeepSeek-R1-Distill-Llama-8B (from left to right).
%\yafu{Consider removing the x/y label (correctness) and explain it in the caption.}
}
\label{fig:quadrant}
\end{minipage}

\section{When Scaling Reasoning Meets Losing Control}
\label{sec:analysis}

As discussed in Section~\ref{sec:exp_res}, there may exist a trade-off between the instruction-following ability and the mathematical reasoning capability of LRMs. In this section, we further investigate this trade-off through a fine-grained error analysis (Section~\ref{sec:ana_error}), examine the effects of different reasoning-oriented training paradigms (Section~\ref{sec:ana_train}), and explore how CoT length impacts reasoning and instruction-following by applying both inference-time and training-aware interventions (Section~\ref{sec:ana_length}).

\subsection{The Intelligence–Obedience Trade-off}
\label{sec:ana_error}

\paragraph{Dilemma between Reasoning and Instruction Following.}
We begin by analyzing the relationship between reasoning and instruction-following through an error-based categorization. Each sample is grouped into one of four categories based on two criteria: (1) whether the math problem was solved correctly, and (2) whether all user-specified constraints were satisfied. The proportions of these four categories are shown in Figure~\ref{fig:quadrant}. We observe that LRMs often struggle to fulfill both objectives simultaneously, as evidenced by the particularly low proportion of (Correct, Followed) cases.
Interestingly, the proportion of (Correct, Followed) is even smaller than that of either (Correct, Unfollowed) or (Incorrect, Followed), suggesting that LRMs frequently sacrifice one objective to achieve the other. In other words, they are more likely to violate constraints when solving problems correctly, and more likely to fail in problem-solving when attempting to follow constraints. This observation is consistent with the trend in Table~\ref{tab:main}.

\begin{wrapfigure}{r}{0.5\textwidth}
\centering
\includegraphics[width=\linewidth]{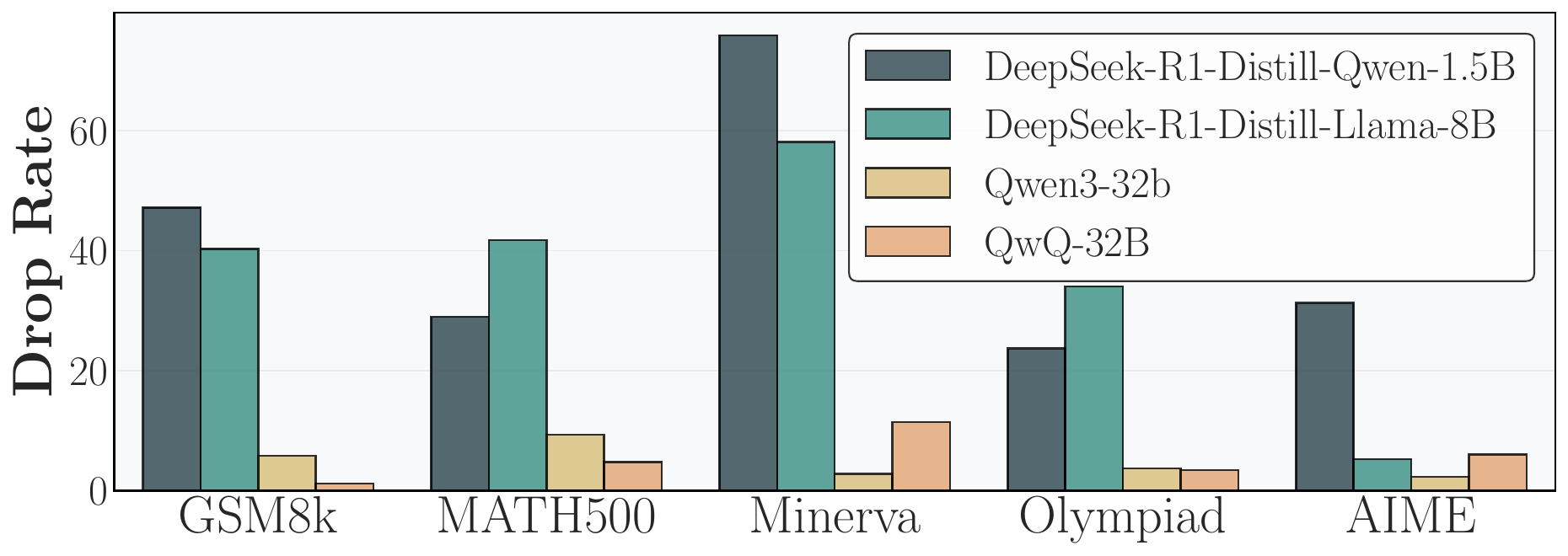}
\caption{Relative correctness drop of four LRMs across five subsets.}
\label{fig:drop_rate}
\end{wrapfigure}

Table~\ref{tab:main} (last column) shows a noticeable degradation in math problem correctness when constraints are introduced. Figure~\ref{fig:drop_rate} further breaks down this effect by dataset. Surprisingly, we find that the drop rate on GSM8K (the easiest subset) is even higher than that on AIME (the hardest), a significantly more challenging benchmark. This suggests that the impact of constraints on reasoning performance is not necessarily correlated with problem difficulty.
In conclusion, the trade-off between instruction-following and reasoning appears to be a general phenomenon across difficulty levels. Notably, LRMs fine-tuned on long CoT traces (e.g., DeepSeek-R1 variants) tend to exhibit more severe performance degradation than RL-trained models like Qwen3-32B and QwQ-32B, possibly due to the inherent limitations of SFT~\cite{chu2025sft}.

%indicating the trade-off between the two abilities. 

% \begin{wraptable}[14]{r}{0.6\linewidth}
%  \includegraphics[width=0.99\linewidth]{figure/drop_rate.pdf}
%  \makeatletter\def\@captype{figure}\makeatother\caption{The drop rate of four LRMs on five subsets.}
%  \label{fig:drop_rate}
% \end{wraptable}

% To investigate whether the intervention only occurs for hard math problems, we measure the relative drop rate $r$ of problem-solving correctness with: $r (m) = \frac{\vert c_w (m)  - c_{w/o} (m) \vert}{c_{w/o}(m)}\times 100\%, $
% and plot the drop rate for each subset in Figure~\ref{fig:drop_rate}, where $m$ is the LRMs to observe and $c_w(\cdot)$ and $c_{w/o}(\cdot)$ denote the correctness of problem-solving with/without the constraint, respectively. 

%A large drop rate thus suggests that the added constraints negatively influence reasoning ability. We plot the drop rate $r$ of DeepSeek-R1-Distill-Qwen-1.5B~\cite{}, DeepSeek-R1-Distill-Llama-8B~\cite{}, QwQ-32B~\cite{} and Qwen3-32b~\cite{} in Figure~\ref{}. 

\begin{wrapfigure}{r}{0.4\textwidth}
  \centering
  \vspace{-25pt}
  \includegraphics[width=\linewidth]{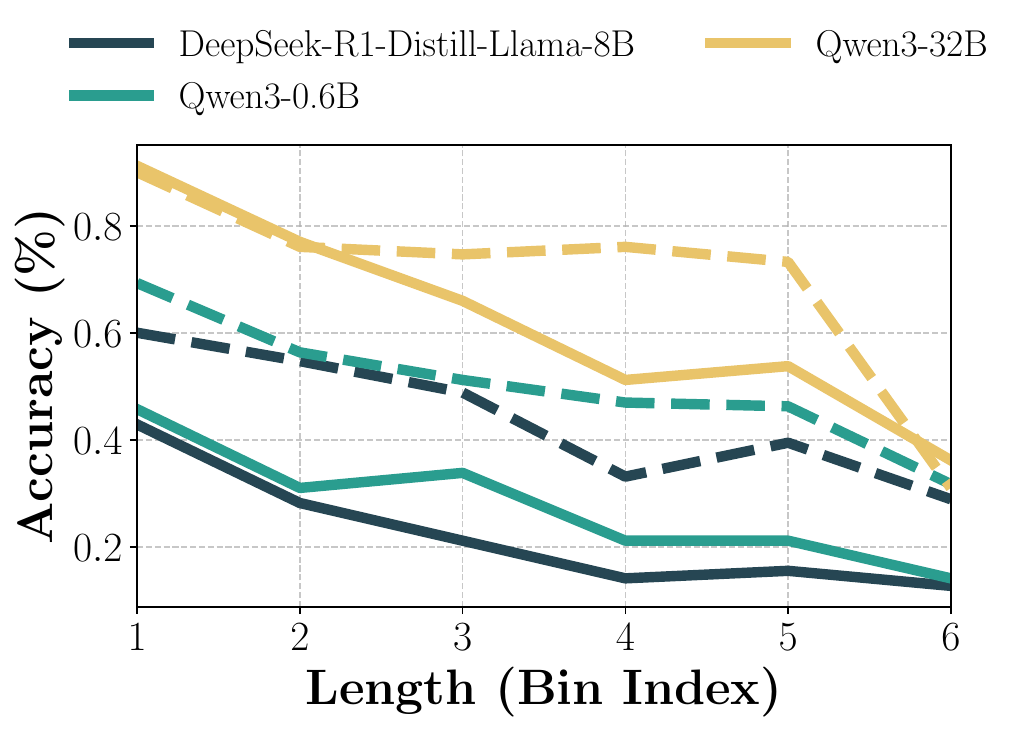}
  \caption{
    HAcc (solid line) and SAcc (dashed line) across six CoT length bins; higher indices correspond to longer CoT generations.
  }
  \label{fig:length_bin}
  \vspace{-10pt}
\end{wrapfigure}

\paragraph{Longer CoTs Impair Instruction Following.}  
We further analyze the impact of CoT length on instruction-following performance. Specifically, for each LRM, we divide its benchmark outputs into six bins based on the number of tokens between the \texttt{<think>} and \texttt{</think>} delimiters. The resulting trends are shown in Figure~\ref{fig:length_bin}. 
Across all three models—DeepSeek-R1-Distill-Llama-8B, Qwen3-0.6B, and Qwen3-32B—we observe a consistent decline in both hard accuracy and soft accuracy as CoT length increases, suggesting a negative correlation between generation length and instruction compliance.
One possible explanation is that longer CoTs, while beneficial for reasoning, increase the distance between the user-specified constraint and the final answer. 
This may dilute the model's attention to the constraint, making accurate instruction-following more difficult (see Section~\ref{sec:ana_length}).

\subsection{How Does Reasoning-Oriented Training Affect Instruction-Following?}
\label{sec:ana_train}

Motivated by the patterns observed in Figure~\ref{fig:drop_rate}, we further investigate how different reasoning-oriented training paradigms affect a model’s instruction-following behavior. Specifically, we examine three representative strategies: (1) \textbf{SFT-only}, (2) \textbf{SFT followed by RL} (SFT+RL), and (3) \textbf{cold-start RL}, which bypasses SFT entirely.

\paragraph{Training Setup.}
We base our experiments on the DeepScaler dataset~\cite{deepscaler2025},
% \footnote{\scriptsize{\url{https://huggingface.co/datasets/agentica-org/DeepScaleR-Preview-Dataset}}}
 which contains approximately 40k math reasoning samples. 
All training is conducted using 16 NVIDIA H100 GPUs.
For SFT-only and SFT+RL settings, we first distill long CoT reasoning traces from QwQ-32B~\cite{qwq32b}, filtering out samples where QwQ-32B fails to generate a correct answer or the CoT exceeds 8192 tokens. This results in 18k high-quality examples. We use models from the Qwen-2.5 and Qwen-2.5-Math series as our base. Since some models are limited to 4096 position embeddings, we extend the RoPE~\cite{su2024roformer} scaling factor $\theta$ from 10,000 to 20,000 to accommodate longer sequences, following prior work~\cite{luffy}.
For reinforcement learning, we adopt the GRPO~\cite{shao2024deepseekmath} framework and use verifiable outcome-based rewards. In addition to standard correctness rewards, we design a format-aware reward variant that grants 0.1 if the model includes special reasoning tokens (e.g., \texttt{<think>} and \texttt{</think>}) and 1.0 for a correct solution.
Details can be referred to in the Appendix. 

\begin{wraptable}{r}{0.6\textwidth}
  \centering
  \small
  \setlength{\tabcolsep}{8pt}
  \renewcommand{\arraystretch}{1.05}
  % \caption{Comparison of different reasoning‐oriented training strategies. Avg. Acc. denotes average performance over all benchmarks. Red and green colors denote increased and decreased performance compared to the model before training. }
  \vspace{-12pt}
  \caption{Comparison of reasoning‐oriented training strategies. Avg.~Acc.\ denotes the average performance across all benchmarks. Cells shaded in 
\colorbox{green!20}{green} and 
\colorbox{red!20}{red} indicate increased and decreased performance, respectively, relative to the base model.}
  \label{tab:qwen_comparison}
  \rowcolors{2}{gray!10}{}% light row shading for readability
  \setlength{\aboverulesep}{0pt}
  \setlength{\belowrulesep}{0pt}
  % colors for increase/decrease
  \definecolor{inc}{RGB}{220,240,220}
  \definecolor{dec}{RGB}{240,220,220}
  \begin{tabular}{lccc}
    \toprule
    \textbf{Model} & \textbf{HAcc} & \textbf{SAcc} & \textbf{Avg.\ Acc.} \\
    \midrule
    \textbf{Qwen2.5-1.5B}       & 10.00       & 27.26       &  1.21 \\
    \quad +SFT                  & \cellcolor{dec}7.86  & \cellcolor{dec}22.70  & \cellcolor{inc}4.20  \\
    \quad +SFT+RL             & \cellcolor{dec}7.86  & \cellcolor{dec}20.44  & \cellcolor{inc}12.54  \\
    \quad +cold‐RL              & \cellcolor{dec}9.52  & \cellcolor{dec}23.97  & \cellcolor{inc}14.58 \\
    \quad\quad w/ format reward & \cellcolor{inc}10.95 & \cellcolor{inc}28.49 & \cellcolor{inc}11.17 \\
    \addlinespace
    \textbf{Qwen-2.5-7B}        & 15.95       & 33.13       & 13.59 \\
    \quad +SFT                  & \cellcolor{dec}7.86  & \cellcolor{dec}21.03  & \cellcolor{inc}23.10 \\
    \quad +SFT+RL             & \cellcolor{dec}7.62  & \cellcolor{dec}21.07  & \cellcolor{inc}32.82 \\
    \quad +cold‐RL              & \cellcolor{dec}10.48 & \cellcolor{dec}27.26  & \cellcolor{inc}28.39 \\
    \quad\quad w/ format reward & \cellcolor{dec}14.52 & \cellcolor{dec}32.50  & \cellcolor{inc}24.80 \\
    \addlinespace
    \textbf{Qwen2.5-Math-1.5B}  &  9.28       & 23.33       & 18.91 \\
    \quad +SFT                  & \cellcolor{dec}7.86  & \cellcolor{dec}21.03  & \cellcolor{dec}14.39 \\
    \quad +SFT+RL             & \cellcolor{dec}7.14  & \cellcolor{dec}20.56  & \cellcolor{inc}24.71 \\
    \quad +cold‐RL              & \cellcolor{dec}8.33  & \cellcolor{dec}21.31  & \cellcolor{inc}24.88 \\
    \quad\quad w/ format reward & \cellcolor{dec}7.62  & \cellcolor{dec}20.08  & \cellcolor{inc}23.95 \\
    \addlinespace
    \textbf{Qwen2.5-Math-7B}    &  9.76       & 23.53       & 20.68 \\
    \quad +SFT                  & \cellcolor{dec}8.09  & \cellcolor{dec}22.06  & \cellcolor{inc}29.11 \\
    \quad +SFT+RL             & \cellcolor{dec}8.57  & \cellcolor{dec}21.03  & \cellcolor{inc}40.65 \\
    \quad +cold‐RL              & \cellcolor{dec}7.85  & \cellcolor{dec}22.62  & \cellcolor{inc}32.61 \\
    \quad\quad w/ format reward & \cellcolor{dec}7.86  & \cellcolor{dec}21.79  & \cellcolor{inc}32.66 \\
    \bottomrule
  \end{tabular}
  \vspace{-15pt}
\end{wraptable}

\paragraph{The Double-Edged Sword of Reasoning-Oriented Training.}
Table~\ref{tab:qwen_comparison} presents the results for different training pathways. 
Avg. Acc. denotes the average benchmark performance with details shown in the Appendix. 
While both SFT and RL contribute to improved reasoning performance, neither of the training strategies enhances instruction-following. On the contrary, we observe a slight but consistent drop in both HAcc and SAcc across the board. The format-aware reward provides a minor improvement on Qwen-2.5-{1.5B, 7B}, but has a negligible effect on the Qwen-2.5-Math series.
These findings reveal a double-edged nature of reasoning-oriented training: while it sharpens the model’s problem-solving ability, it simultaneously dulls its capacity to follow user instructions. 
This underscores a central \textbf{dilemma} in current training paradigms: \textit{enhancing intelligence often comes at the expense of obedience}.

\subsection{How does the CoT Length Affect Instruction Following?}
\label{sec:ana_length}
Prior work has shown that generating longer chains of thought (CoT) can significantly improve a model’s ability to solve complex tasks~\cite{jin2024impact}. 
However, as illustrated in Figure~\ref{fig:length_bin}, we observe that scaling up CoT length can come at the cost of instruction-following accuracy. 
To better understand this phenomenon, we systematically investigate how CoT length influences instruction adherence.

\begin{wrapfigure}{r}{0.4\textwidth}
  \centering
  \vspace{-5pt}
  \includegraphics[width=\linewidth]{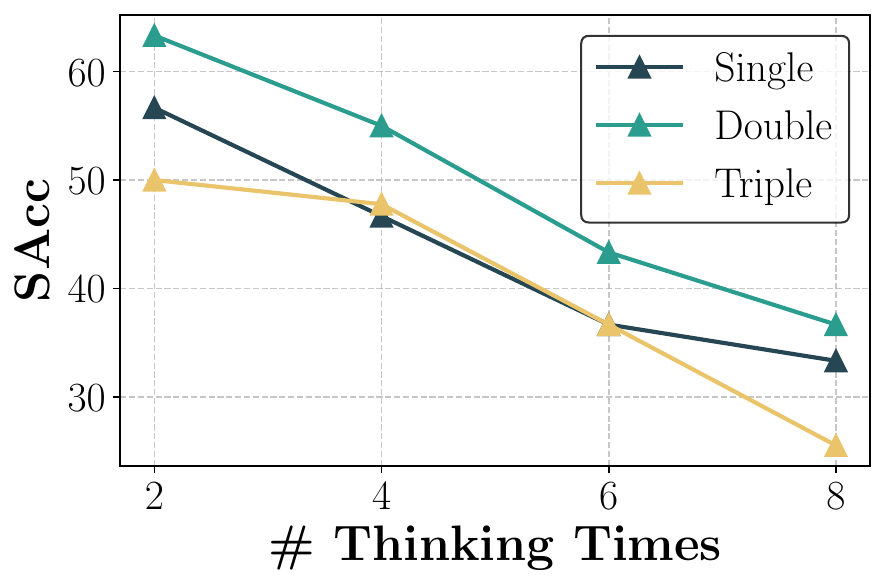}
  % \vspace{-10pt}
  \caption{The trend of SAcc variation on GSM8K subset as the number of budget forcing increases from 2 to 8.}
  \label{fig:gsm8k_ignore}
  % \vspace{-20pt}
\end{wrapfigure}

\paragraph{The More Thinking, the Less Following.} 
We begin by artificially increasing the CoT length using budget forcing~\cite{2025s1}, which appends the token "Wait" each time the model attempts to terminate the reasoning process. 
This encourages the model to continue generating longer CoTs. 
The experiment is performed on DeepSeek-R1-Distill-Qwen-1.5B, and Figure~\ref{fig:gsm8k_ignore} shows the instruction-following performance as the number of budget-forcing steps $N$ increases from 2 to 8.
The results reveal a clear trend: SAcc steadily declines as CoT length increases, suggesting that excessively long CoTs may impair the model’s ability to follow instructions. 
This degradation likely stems from the increasing distance between the instruction and the final output, which may dilute the model's attention to user-specified constraints.

\paragraph{Controlling CoT Length During RL Training.}
Beyond inference-time manipulation, we investigate whether controlling the length of CoT during reinforcement learning has a similar impact on instruction-following. Specifically, we continue RL training on DeepSeek-R1-Distill-Qwen-1.5B using the DeepScaler dataset~\cite{deepscaler2025}, varying the maximum response length during rollouts.

% \begin{wraptable}{r}{0.48\textwidth}
%   \centering
%   % \footnotesize
%   \caption{Impact of varying roll‐out sizes.}
%   \label{tab:rollout_effect}
%    \begin{small}
%   \setlength{\tabcolsep}{3pt}
%   % \renewcommand{\arraystretch}{1}
%   \begin{tabular}{lccc}
%     \toprule
%     \textbf{Model}                     & \textbf{HAcc} & \textbf{SAcc} & \textbf{Math} \\
%     \midrule
%     Original                           & 17.14 & 36.62 & 36.13 \\
%     \midrule
%     \quad +cold‐RL (1k)                & 19.05 & 39.88 & 28.73 \\
%     \quad +cold‐RL (2k)                & 16.43 & 36.75 & 36.32 \\
%     \quad +cold‐RL (4k)                & 16.91 & 35.87 & 40.03 \\
%     \quad +cold‐RL (8k)                & 14.29 & 34.13 & 39.82 \\
%     \bottomrule
%   \end{tabular}
%   \end{small}
% \end{wraptable}

% \begin{wraptable}{r}{0.4\textwidth}
%   \centering
%   \footnotesize
%    \caption{Impact of varying roll‐out sizes.}
%    \label{tab:rollout_effect}
%   \setlength{\tabcolsep}{2.5pt}
%   % \renewcommand{\arraystretch}{0.8}
%  \begin{small}
  
%   \begin{tabular}{lccc}
%     \toprule
%     \textbf{Model}                     & \textbf{HAcc} & \textbf{SAcc} & \textbf{Math} \\
%     \midrule
%     Original                           & 17.14 & 36.62 & 36.13 \\
%     \midrule
%     \quad +cold‐RL (1k)                & 19.05 & 39.88 & 28.73 \\
%     \quad +cold‐RL (2k)                & 16.43 & 36.75 & 36.32 \\
%     \quad +cold‐RL (4k)                & 16.91 & 35.87 & 40.03 \\
%     \quad +cold‐RL (8k)                & 14.29 & 34.13 & 39.82 \\
%     \bottomrule
%   \end{tabular}
%   \end{small}
% \end{wraptable}

\begin{wraptable}{r}{0.5\textwidth}
  \centering
  \footnotesize
  \caption{Impact of the maximum response length during RL. Cells shaded in red denote lower performance relative to the base (\textit{Original}), with intensity proportional to the drop magnitude.}
  \label{tab:rollout_effect}
  \setlength{\tabcolsep}{5pt}
  \begin{small}
  \rowcolors{2}{gray!10}{}      % alternate row shading
  \begin{tabular}{lccc}
    \toprule
    \textbf{Model}                     & \textbf{HAcc}       & \textbf{SAcc}       & \textbf{Avg. Acc.} \\
    \midrule
    Original                           & 17.14               & 36.62               & 36.13 \\
    \midrule
    \quad +cold‐RL (1k)                & 19.05               & 39.88               & \cellcolor{red!70}28.73 \\
    \quad +cold‐RL (2k)                & \cellcolor{red!10}16.43 & 36.75           & 36.32 \\
    \quad +cold‐RL (4k)                & \cellcolor{red!5}16.91  & \cellcolor{red!10}35.87 & 40.03 \\
    \quad +cold‐RL (8k)                & \cellcolor{red!30}14.29 & \cellcolor{red!25}34.13 & 39.82 \\
    \bottomrule
  \end{tabular}
  \end{small}
\end{wraptable}

In this setup, overlong responses are truncated and receive no outcome reward, encouraging the model to learn more concise reasoning traces. 
We adopt a pure outcome-based reward function and conduct RL training for three epochs, varying the maximum rollout length from 1k to 8k tokens.
The results, shown in Table~\ref{tab:rollout_effect}, reveal a clear trend: as the maximum rollout length increases, math reasoning performance (averaged across AIME2024, AIME2025, AMC2023, Minerva, and Olympiad) improves, while both hard accuracy and soft accuracy consistently decline. 
This observation further reinforces our conclusion: \textit{ reasoning-oriented training that \textbf{favors longer CoTs} can inadvertently \textbf{harm} \textbf{instruction-following} fidelity, highlighting a persistent trade-off between reasoning strength and obedience to user constraints.}

\begin{wraptable}{r}{0.6\textwidth}
  \centering
  \footnotesize
   \caption{Effect of \texttt{+repeat} on model performance.}
  \setlength{\tabcolsep}{2pt}
 \begin{small}
  \label{tab:repeat_effect}
  \begin{tabular}{lccc}
    \toprule
    \textbf{Model} & \textbf{HAcc} & \textbf{SAcc} & \textbf{Correctness} \\
    \midrule
    DeepSeek-R1-Distill-Qwen-1.5B & 17.14 & 36.62 & 31.67 \\
    \quad \texttt{+repeat}                 & 21.66 & 42.58 & 22.38 \\
    \addlinespace
    Open-Reasoner-Zero-7B         & 13.57 & 32.26 & 51.90 \\
    \quad \texttt{+repeat}           & 14.53 & 33.14 & 30.00 \\
    \addlinespace
    Qwen3-32B                     & 43.81 & 62.82 & 70.00 \\
    \quad \texttt{+repeat}                 & 59.29 & 68.34 & 63.81 \\
    \bottomrule
  \end{tabular}
  \end{small}
\end{wraptable}

\paragraph{Bringing Instructions Closer Improves Obedience at the Cost of Intelligence.}
One possible explanation for the negative impact of lengthy CoTs on instruction-following is that extended reasoning increases the distance between the user query and the final answer, making it more likely for the model to overlook the original constraint. 
To preliminarily verify this hypothesis, we propose a simple yet effective remedy: repeating the constraint at the end of the CoT.
Concretely, we manually append the token "Wait" to prolong the CoT and then \textbf{reintroduce} the original constraint immediately afterward. 
As a result, the constraint appears twice in the input, i.e., once before the CoT begins and once again at the end, thereby shortening its contextual distance from the final answer.
Experimental results on DeepSeek-R1-Distill-Qwen-1.5B, Open-Reasoner-Zero-7B, and Qwen3-32B are shown in Table~\ref{tab:repeat_effect}. This straightforward intervention leads to clear improvements in instruction-following (SAcc and HAcc), albeit at a modest cost to problem-solving accuracy.
These findings further confirm the inherent trade-off between reasoning depth and obedience during inference: \textit{enhancing one often comes at the expense of the other.}

\section{Conclusion}

While large reasoning models continue to demonstrate impressive progress in mathematical problem-solving, our study reveals a persistent and underexplored trade-off between reasoning strength and instruction-following fidelity.

Through \benchmarkname, a benchmark tailored for evaluating instruction adherence in math reasoning tasks, we show that reasoning scaling does not guarantee control. Empirical results reveal that longer chains of thought and reasoning-oriented training methods (e.g., SFT and RL) often impair a model’s ability to comply with user-specified constraints.
These findings highlight a core tension in the development of LRMs: as models become more intelligent, they often become less controllable. This dilemma is central to the alignment problem in reasoning-centric systems. Addressing it requires rethinking current training paradigms to build models that can reason effectively without drifting from user intent.
We hope that our benchmark and findings serve as a foundation for future research that bridges the growing gap between intelligence and obedience in large reasoning models.

\newpage

\bibliography{neurips_2025.bib}

\begin{thebibliography}{10}

\bibitem{qu2025survey}
Xiaoye Qu, Yafu Li, Zhaochen Su, Weigao Sun, Jianhao Yan, Dongrui Liu, Ganqu Cui, Daizong Liu, Shuxian Liang, Junxian He, et~al.
\newblock A survey of efficient reasoning for large reasoning models: Language, multimodality, and beyond.
\newblock {\em arXiv preprint arXiv:2503.21614}, 2025.

\bibitem{o3}
OpenAI.
\newblock Introducing openai o3 and o4-mini.
\newblock \url{https://openai.com/index/introducing-o3-and-o4-mini/}.

\bibitem{deepseek-r1}
DeepSeek-AI.
\newblock Deepseek-r1: Incentivizing reasoning capability in llms via reinforcement learning, 2025.

\bibitem{kimi}
Kimi Team, Angang Du, Bofei Gao, Bowei Xing, Changjiu Jiang, Cheng Chen, Cheng Li, Chenjun Xiao, Chenzhuang Du, Chonghua Liao, Chuning Tang, Congcong Wang, Dehao Zhang, Enming Yuan, Enzhe Lu, Fengxiang Tang, Flood Sung, Guangda Wei, Guokun Lai, Haiqing Guo, Han Zhu, Hao Ding, Hao Hu, Hao Yang, Hao Zhang, Haotian Yao, Haotian Zhao, Haoyu Lu, Haoze Li, Haozhen Yu, Hongcheng Gao, Huabin Zheng, Huan Yuan, Jia Chen, Jianhang Guo, Jianlin Su, Jianzhou Wang, Jie Zhao, Jin Zhang, Jingyuan Liu, Junjie Yan, Junyan Wu, Lidong Shi, Ling Ye, Longhui Yu, Mengnan Dong, Neo Zhang, Ningchen Ma, Qiwei Pan, Qucheng Gong, Shaowei Liu, Shengling Ma, Shupeng Wei, Sihan Cao, Siying Huang, Tao Jiang, Weihao Gao, Weimin Xiong, Weiran He, Weixiao Huang, Wenhao Wu, Wenyang He, Xianghui Wei, Xianqing Jia, Xingzhe Wu, Xinran Xu, Xinxing Zu, Xinyu Zhou, Xuehai Pan, Y.~Charles, Yang Li, Yangyang Hu, Yangyang Liu, Yanru Chen, Yejie Wang, Yibo Liu, Yidao Qin, Yifeng Liu, Ying Yang, Yiping Bao, Yulun Du, Yuxin Wu, Yuzhi Wang, Zaida Zhou,
  Zhaoji Wang, Zhaowei Li, Zhen Zhu, Zheng Zhang, Zhexu Wang, Zhilin Yang, Zhiqi Huang, Zihao Huang, Ziyao Xu, and Zonghan Yang.
\newblock Kimi k1.5: Scaling reinforcement learning with llms, 2025.

\bibitem{OlympiadBench}
Chaoqun He, Renjie Luo, Yuzhuo Bai, Shengding Hu, Zhen~Leng Thai, Junhao Shen, Jinyi Hu, Xu~Han, Yujie Huang, Yuxiang Zhang, Jie Liu, Lei Qi, Zhiyuan Liu, and Maosong Sun.
\newblock Olympiadbench: A challenging benchmark for promoting agi with olympiad-level bilingual multimodal scientific problems, 2024.

\bibitem{math500}
Dan Hendrycks, Collin Burns, Saurav Kadavath, Akul Arora, Steven Basart, Eric Tang, Dawn Song, and Jacob Steinhardt.
\newblock Measuring mathematical problem solving with the math dataset.
\newblock {\em arXiv preprint arXiv:2103.03874}, 2021.

\bibitem{aime_1983_2024}
Hemish Veeraboina.
\newblock Aime problem set 1983-2024, 2023.

\bibitem{deepseek-math-prover}
Z.~Z. Ren, Zhihong Shao, Junxiao Song, Huajian Xin, Haocheng Wang, Wanjia Zhao, Liyue Zhang, Zhe Fu, Qihao Zhu, Dejian Yang, Z.~F. Wu, Zhibin Gou, Shirong Ma, Hongxuan Tang, Yuxuan Liu, Wenjun Gao, Daya Guo, and Chong Ruan.
\newblock Deepseek-prover-v2: Advancing formal mathematical reasoning via reinforcement learning for subgoal decomposition, 2025.

\bibitem{cot}
Jason Wei, Xuezhi Wang, Dale Schuurmans, Maarten Bosma, Fei Xia, Ed~Chi, Quoc~V Le, Denny Zhou, et~al.
\newblock Chain-of-thought prompting elicits reasoning in large language models.
\newblock {\em Advances in neural information processing systems}, 35:24824--24837, 2022.

\bibitem{rl-vr}
Yi~Su, Dian Yu, Linfeng Song, Juntao Li, Haitao Mi, Zhaopeng Tu, Min Zhang, and Dong Yu.
\newblock Crossing the reward bridge: Expanding rl with verifiable rewards across diverse domains, 2025.

\bibitem{surveyllmasajudge}
Jiawei Gu, Xuhui Jiang, Zhichao Shi, Hexiang Tan, Xuehao Zhai, Chengjin Xu, Wei Li, Yinghan Shen, Shengjie Ma, Honghao Liu, Saizhuo Wang, Kun Zhang, Yuanzhuo Wang, Wen Gao, Lionel Ni, and Jian Guo.
\newblock A survey on llm-as-a-judge, 2025.

\bibitem{ifeval}
Jeffrey Zhou, Tianjian Lu, Swaroop Mishra, Siddhartha Brahma, Sujoy Basu, Yi~Luan, Denny Zhou, and Le~Hou.
\newblock Instruction-following evaluation for large language models.
\newblock {\em arXiv preprint arXiv:2311.07911}, 2023.

\bibitem{followbench}
Yuxin Jiang, Yufei Wang, Xingshan Zeng, Wanjun Zhong, Liangyou Li, Fei Mi, Lifeng Shang, Xin Jiang, Qun Liu, and Wei Wang.
\newblock Followbench: A multi-level fine-grained constraints following benchmark for large language models.
\newblock {\em arXiv preprint arXiv:2310.20410}, 2023.

\bibitem{2025s1}
Niklas Muennighoff, Zitong Yang, Weijia Shi, Xiang~Lisa Li, Li~Fei-Fei, Hannaneh Hajishirzi, Luke Zettlemoyer, Percy Liang, Emmanuel Cand{\`e}s, and Tatsunori Hashimoto.
\newblock s1: Simple test-time scaling.
\newblock {\em arXiv preprint arXiv:2501.19393}, 2025.

\bibitem{ye2025limoreasoning}
Yixin Ye, Zhen Huang, Yang Xiao, Ethan Chern, Shijie Xia, and Pengfei Liu.
\newblock Limo: Less is more for reasoning, 2025.

\bibitem{yeo2025demystifying}
Edward Yeo, Yuxuan Tong, Morry Niu, Graham Neubig, and Xiang Yue.
\newblock Demystifying long chain-of-thought reasoning in llms, 2025.

\bibitem{chu2025sft}
Tianzhe Chu, Yuexiang Zhai, Jihan Yang, Shengbang Tong, Saining Xie, Sergey Levine, and Yi~Ma.
\newblock {SFT} memorizes, {RL} generalizes: A comparative study of foundation model post-training.
\newblock In {\em The Second Conference on Parsimony and Learning (Recent Spotlight Track)}, 2025.

\bibitem{liu2025oatzero}
Zichen Liu, Changyu Chen, Wenjun Li, Tianyu Pang, Chao Du, and Min Lin.
\newblock There may not be aha moment in r1-zero-like training — a pilot study.
\newblock \url{https://oatllm.notion.site/oat-zero}, 2025.
\newblock Notion Blog.

\bibitem{yu2025dapo}
Qiying Yu, Zheng Zhang, Ruofei Zhu, Yufeng Yuan, Xiaochen Zuo, Yu~Yue, Tiantian Fan, Gaohong Liu, Lingjun Liu, Xin Liu, et~al.
\newblock Dapo: An open-source llm reinforcement learning system at scale.
\newblock {\em arXiv preprint arXiv:2503.14476}, 2025.

\bibitem{prime}
Ganqu Cui, Lifan Yuan, Zefan Wang, Hanbin Wang, Wendi Li, Bingxiang He, Yuchen Fan, Tianyu Yu, Qixin Xu, Weize Chen, et~al.
\newblock Process reinforcement through implicit rewards.
\newblock {\em arXiv preprint arXiv:2502.01456}, 2025.

\bibitem{luffy}
Jianhao Yan, Yafu Li, Zican Hu, Zhi Wang, Ganqu Cui, Xiaoye Qu, Yu~Cheng, and Yue Zhang.
\newblock Learning to reason under off-policy guidance, 2025.

\bibitem{yang2025thinking}
Wang Yang, Hongye Jin, Jingfeng Yang, Vipin Chaudhary, and Xiaotian Han.
\newblock Thinking preference optimization.
\newblock {\em arXiv preprint arXiv:2502.13173}, 2025.

\bibitem{dubois2023alpacafarm}
Yann Dubois, Xuechen Li, Rohan Taori, Tianyi Zhang, Ishaan Gulrajani, Jimmy Ba, Carlos Guestrin, Percy Liang, and Tatsunori~B. Hashimoto.
\newblock Alpacafarm: A simulation framework for methods that learn from human feedback, 2023.

\bibitem{chiang2023vicuna}
Wei-Lin Chiang, Zhuohan Li, Zi~Lin, Ying Sheng, Zhanghao Wu, Hao Zhang, Lianmin Zheng, Siyuan Zhuang, Yonghao Zhuang, Joseph~E. Gonzalez, Ion Stoica, and Eric~P. Xing.
\newblock Vicuna: An open-source chatbot impressing gpt-4 with 90\%* chatgpt quality, March 2023.

\bibitem{xia2024fofo}
Congying Xia, Chen Xing, Jiangshu Du, Xinyi Yang, Yihao Feng, Ran Xu, Wenpeng Yin, and Caiming Xiong.
\newblock {FOFO}: A benchmark to evaluate {LLM}s' format-following capability.
\newblock In Lun-Wei Ku, Andre Martins, and Vivek Srikumar, editors, {\em Proceedings of the 62nd Annual Meeting of the Association for Computational Linguistics (Volume 1: Long Papers)}, pages 680--699, Bangkok, Thailand, August 2024. Association for Computational Linguistics.

\bibitem{tang2024struc}
Xiangru Tang, Yiming Zong, Jason Phang, Yilun Zhao, Wangchunshu Zhou, Arman Cohan, and Mark Gerstein.
\newblock Struc-bench: Are large language models good at generating complex structured tabular data?
\newblock In Kevin Duh, Helena Gomez, and Steven Bethard, editors, {\em Proceedings of the 2024 Conference of the North American Chapter of the Association for Computational Linguistics: Human Language Technologies (Volume 2: Short Papers)}, pages 12--34, Mexico City, Mexico, June 2024. Association for Computational Linguistics.

\bibitem{he2024multi}
Yun He, Di~Jin, Chaoqi Wang, Chloe Bi, Karishma Mandyam, Hejia Zhang, Chen Zhu, Ning Li, Tengyu Xu, Hongjiang Lv, et~al.
\newblock Multi-if: Benchmarking llms on multi-turn and multilingual instructions following.
\newblock {\em arXiv preprint arXiv:2410.15553}, 2024.

\bibitem{li2025structflowbench}
Jinnan Li, Jinzhe Li, Yue Wang, Yi~Chang, and Yuan Wu.
\newblock Structflowbench: A structured flow benchmark for multi-turn instruction following.
\newblock {\em arXiv preprint arXiv:2502.14494}, 2025.

\bibitem{han2025can}
Chi Han.
\newblock Can language models follow multiple turns of entangled instructions?
\newblock {\em arXiv preprint arXiv:2503.13222}, 2025.

\bibitem{sirdeshmukh2025multichallenge}
Ved Sirdeshmukh, Kaustubh Deshpande, Johannes Mols, Lifeng Jin, Ed-Yeremai Cardona, Dean Lee, Jeremy Kritz, Willow Primack, Summer Yue, and Chen Xing.
\newblock Multichallenge: A realistic multi-turn conversation evaluation benchmark challenging to frontier llms.
\newblock {\em arXiv preprint arXiv:2501.17399}, 2025.

\bibitem{wu2024lifbench}
Xiaodong Wu, Minhao Wang, Yichen Liu, Xiaoming Shi, He~Yan, Xiangju Lu, Junmin Zhu, and Wei Zhang.
\newblock Lifbench: Evaluating the instruction following performance and stability of large language models in long-context scenarios.
\newblock {\em arXiv preprint arXiv:2411.07037}, 2024.

\bibitem{li2025xifbench}
Zhenyu Li, Kehai Chen, Yunfei Long, Xuefeng Bai, Yaoyin Zhang, Xuchen Wei, Juntao Li, and Min Zhang.
\newblock Xifbench: Evaluating large language models on multilingual instruction following.
\newblock {\em arXiv preprint arXiv:2503.07539}, 2025.

\bibitem{zhang2025iheval}
Zhihan Zhang, Shiyang Li, Zixuan Zhang, Xin Liu, Haoming Jiang, Xianfeng Tang, Yifan Gao, Zheng Li, Haodong Wang, Zhaoxuan Tan, Yichuan Li, Qingyu Yin, Bing Yin, and Meng Jiang.
\newblock {IHE}val: Evaluating language models on following the instruction hierarchy.
\newblock In Luis Chiruzzo, Alan Ritter, and Lu~Wang, editors, {\em Proceedings of the 2025 Conference of the Nations of the Americas Chapter of the Association for Computational Linguistics: Human Language Technologies (Volume 1: Long Papers)}, pages 8374--8398, Albuquerque, New Mexico, April 2025. Association for Computational Linguistics.

\bibitem{hayati2025chain}
Shirley~Anugrah Hayati, Taehee Jung, Tristan Bodding-Long, Sudipta Kar, Abhinav Sethy, Joo-Kyung Kim, and Dongyeop Kang.
\newblock Chain-of-instructions: Compositional instruction tuning on large language models.
\newblock In {\em Proceedings of the AAAI Conference on Artificial Intelligence}, volume~39, pages 24005--24013, 2025.

\bibitem{yan-etal-2024-refutebench}
Jianhao Yan, Yun Luo, and Yue Zhang.
\newblock {R}efute{B}ench: Evaluating refuting instruction-following for large language models.
\newblock In Lun-Wei Ku, Andre Martins, and Vivek Srikumar, editors, {\em Findings of the Association for Computational Linguistics: ACL 2024}, pages 13775--13791, Bangkok, Thailand, August 2024. Association for Computational Linguistics.

\bibitem{yan2025refutebench20agentic}
Jianhao Yan, Yun Luo, and Yue Zhang.
\newblock Refutebench 2.0 -- agentic benchmark for dynamic evaluation of llm responses to refutation instruction, 2025.

\bibitem{complexbench}
Bosi Wen, Pei Ke, Xiaotao Gu, Lindong Wu, Hao Huang, Jinfeng Zhou, Wenchuang Li, Binxin Hu, Wendy Gao, Jiaxing Xu, et~al.
\newblock Benchmarking complex instruction-following with multiple constraints composition.
\newblock {\em Advances in Neural Information Processing Systems}, 37:137610--137645, 2024.

\bibitem{hurst2024gpt}
Aaron Hurst, Adam Lerer, Adam~P Goucher, Adam Perelman, Aditya Ramesh, Aidan Clark, AJ~Ostrow, Akila Welihinda, Alan Hayes, Alec Radford, et~al.
\newblock Gpt-4o system card.
\newblock {\em arXiv preprint arXiv:2410.21276}, 2024.

\bibitem{gsm8k}
Karl Cobbe, Vineet Kosaraju, Mohammad Bavarian, Mark Chen, Heewoo Jun, Lukasz Kaiser, Matthias Plappert, Jerry Tworek, Jacob Hilton, Reiichiro Nakano, et~al.
\newblock Training verifiers to solve math word problems.
\newblock {\em arXiv preprint arXiv:2110.14168}, 2021.

\bibitem{de2013minerva}
Arup De, Maya Gokhale, Rajesh Gupta, and Steven Swanson.
\newblock Minerva: Accelerating data analysis in next-generation ssds.
\newblock In {\em 2013 IEEE 21st Annual International Symposium on Field-Programmable Custom Computing Machines}, pages 9--16. IEEE, 2013.

\bibitem{vllm}
Woosuk Kwon, Zhuohan Li, Siyuan Zhuang, Ying Sheng, Lianmin Zheng, Cody~Hao Yu, Joseph~E. Gonzalez, Hao Zhang, and Ion Stoica.
\newblock Efficient memory management for large language model serving with pagedattention.
\newblock In {\em Proceedings of the ACM SIGOPS 29th Symposium on Operating Systems Principles}, 2023.

\bibitem{qwen3}
Qwen Team.
\newblock Qwen3, April 2025.

\bibitem{simplerl}
Weihao Zeng, Yuzhen Huang, Qian Liu, Wei Liu, Keqing He, Zejun Ma, and Junxian He.
\newblock Simplerl-zoo: Investigating and taming zero reinforcement learning for open base models in the wild.
\newblock {\em arXiv preprint arXiv:2503.18892}, 2025.

\bibitem{qwen-math}
An~Yang, Beichen Zhang, Binyuan Hui, Bofei Gao, Bowen Yu, Chengpeng Li, Dayiheng Liu, Jianhong Tu, Jingren Zhou, Junyang Lin, Keming Lu, Mingfeng Xue, Runji Lin, Tianyu Liu, Xingzhang Ren, and Zhenru Zhang.
\newblock Qwen2.5-math technical report: Toward mathematical expert model via self-improvement.
\newblock {\em arXiv preprint arXiv:2409.12122}, 2024.

\bibitem{deepscaler2025}
Michael Luo, Sijun Tan, Justin Wong, Xiaoxiang Shi, William~Y. Tang, Manan Roongta, Colin Cai, Jeffrey Luo, Li~Erran Li, Raluca~Ada Popa, and Ion Stoica.
\newblock Deepscaler: Surpassing o1-preview with a 1.5b model by scaling rl, 2025.
\newblock Notion Blog.

\bibitem{l1}
Pranjal Aggarwal and Sean Welleck.
\newblock L1: Controlling how long a reasoning model thinks with reinforcement learning.
\newblock {\em arXiv preprint arXiv:2503.04697}, 2025.

\bibitem{hu2025open}
Jingcheng Hu, Yinmin Zhang, Qi~Han, Daxin Jiang, Xiangyu Zhang, and Heung-Yeung Shum.
\newblock Open-reasoner-zero: An open source approach to scaling up reinforcement learning on the base model.
\newblock {\em arXiv preprint arXiv:2503.24290}, 2025.

\bibitem{openr1}
Hugging Face.
\newblock Open r1: A fully open reproduction of deepseek-r1, January 2025.

\bibitem{qwq32b}
Qwen Team.
\newblock Qwq-32b: Embracing the power of reinforcement learning, March 2025.

\bibitem{reasoner-zero}
Jingcheng Hu, Yinmin Zhang, Qi~Han, Daxin Jiang, Xiangyu Zhang, and Heung-Yeung Shum.
\newblock Open-reasoner-zero: An open source approach to scaling up reinforcement learning on the base model, 2025.

\bibitem{su2024roformer}
Jianlin Su, Murtadha Ahmed, Yu~Lu, Shengfeng Pan, Wen Bo, and Yunfeng Liu.
\newblock Roformer: Enhanced transformer with rotary position embedding.
\newblock {\em Neurocomputing}, 568:127063, 2024.

\bibitem{shao2024deepseekmath}
Zhihong Shao, Peiyi Wang, Qihao Zhu, Runxin Xu, Junxiao Song, Xiao Bi, Haowei Zhang, Mingchuan Zhang, YK~Li, Y~Wu, et~al.
\newblock Deepseekmath: Pushing the limits of mathematical reasoning in open language models.
\newblock {\em arXiv preprint arXiv:2402.03300}, 2024.

\bibitem{jin2024impact}
Mingyu Jin, Qinkai Yu, Dong Shu, Haiyan Zhao, Wenyue Hua, Yanda Meng, Yongfeng Zhang, and Mengnan Du.
\newblock The impact of reasoning step length on large language models.
\newblock In Lun-Wei Ku, Andre Martins, and Vivek Srikumar, editors, {\em Findings of the Association for Computational Linguistics: ACL 2024}, pages 1830--1842, Bangkok, Thailand, August 2024. Association for Computational Linguistics.

\end{thebibliography}
\bibliographystyle{unsrt}

\appendix
\newpage

% Add appendix heading
% \part*{Appendix}

\section{Overview of the Appendix}
This Appendix is organized as follows:
\begin{itemize}[wide=0.\parindent,noitemsep,topsep=0.em]
    \item Section~\ref{sec:limit} and Section~\ref{sec:ethics} discussed the limitation and ethic considerations of our study, respectively. 
    \item Section \ref{sec:hyper} elaborate on the hyper-parameters used for our reasoning-oriented training in Section 5.2.
    \item Section~\ref{sec:more_benchmark} provides more detailed results on our benchmark to facilitate analysis on the difficulty of math problems and the number of constraints.
    \item Section~\ref{sec:more_math} contains detailed reasoning performance for LRMs trained in Section 5.
    \item Section~\ref{sec:more_if} provides more details for our preliminary observation on existing instruction-following benchmarks. 
    \item Section~\ref{sec:constraint_list} lists the constraints used in our proposed \benchmarkname benchmark.
\end{itemize}

% \addcontentsline{toc}{part}{Appendix}  % Add to main TOC

% % Create a local TOC for the appendix
% \startcontents[appendix]
% % \printcontents[appendix]{}{1}{\section*{Contents of Appendix}}
% \printcontents[appendix]{}{1}{}

\section{Limitations}
\label{sec:limit}
The limitations of this study can be summarized as below:
\begin{itemize}[wide=0.\parindent,noitemsep,topsep=0.em]
\item In this study, we evaluate  $23$ recently released LRMs in text modality, and we plan to leave the benchmarking of large vision reasoning models for future work. 

\item When investigating how reasoning-oriented training affects instruction-following, we mainly use GRPO~\cite{shao2024deepseekmath} for RL training because of its simplicity, stability, and widespread practical adoption. Experimenting with other RL training algorithms is left for future work.  

\end{itemize}

\section{Ethics Considerations}
\label{sec:ethics}
Our proposed \benchmarkname evaluates the instruction-following ability of publicly released LRMs, adhering strictly to the Neurips Code of Ethics. The math problems used for our benchmark are collected from free public datasets, and the construction of our benchmark does not involve recruiting crowdsource workers or human annotators. Our benchmark should only be used for research, not for any malicious purpose.

\section{Hyper-parameter Setting}
\label{sec:hyper}
Our experiments on different reasoning-oriented training strategies in Section 5.2 are conducted on a cloud Linux server with Ubuntu 16.04 operating system. The codes are written in Python 3.10 with the huggingface libraries\footnote{\scriptsize\url{https://github.com/huggingface/transformers}}. We run our experiments on 16 Nvidia H100 with 80GiB GPU memory.
The detailed hyper-parameter settings for supervised fine-tuning and reinforcement learning are shown in Table~\ref{tab:hyper_parameter}, which mostly follow the default setting in VeRL framework~\footnote{\scriptsize\url{https://github.com/volcengine/verl}}.

\begin{minipage}{1.025\textwidth}
\centering
\makeatletter\def\@captype{table}\makeatother\caption{
The value of the hyper-parameters in our reasoning-oriented training experiment (Section 5.2) for SFT (left) and RL (right). }
\begin{minipage}[p!]{0.45\textwidth}
\renewcommand{\arraystretch}{1.05}

\begin{tabular}{lc}
\toprule
\textbf{Hyper-parameter}    & \textbf{Value} \\
\midrule
\texttt{batch\_size}        & 256         \\
\texttt{micro\_batch\_size}  & 1           \\
\texttt{max\_length}        & 8192        \\
\texttt{rope\_theta}        & 20000       \\
\texttt{lr}                 & 1e-6     \\
\texttt{betas}              & (0.9, 0.95) \\
\texttt{weight\_decay}       & 0.01        \\
\texttt{warmup\_ratio}       & 0.1         \\
\texttt{schedule}           & cosine      \\
\texttt{clip\_grad}          & 1           \\
\texttt{epoch}              & 3           \\
\texttt{truncation}         & right       \\
\texttt{sliding\_window}    & none        \\
\bottomrule
\end{tabular}

 \label{tab:hyper_parameter_sft}
\end{minipage}
\begin{minipage}[p!]{0.45\textwidth}
\renewcommand{\arraystretch}{1.0}

\begin{tabular}{lc}
\toprule
\textbf{Hyper-parameter} & \textbf{Value} \\
\midrule
\texttt{max\_prompt\_length}   & 1024     \\
\texttt{max\_response\_length} & 3072     \\
\texttt{lr}                    & 1e-6 \\
\texttt{batch\_size}           & 128      \\
\texttt{mini\_batch\_size}     & 64       \\
\texttt{grad\_clip}            & 1        \\
\texttt{clip\_ratio}           & 0.2      \\
\texttt{entropy\_coeff}        & 0.001    \\
\texttt{kl\_loss\_coef}        & 0.001    \\
\texttt{rl\_epoch}             & 1        \\
\texttt{warmup\_ratio}         & 0        \\
\texttt{schedule}              & constant \\
\texttt{rollout\_n}            & 8        \\
\texttt{rollout\_temperature}  & 1        \\
\bottomrule
\end{tabular}

 \label{tab:hyper_parameter_rl}
\end{minipage}
\label{tab:hyper_parameter}
\end{minipage}

\section{More Benchmark Results}
\label{sec:more_benchmark}
In Section 4.3, we visualize the model performance grouped by the source of math problems and the number of constraints. In this section, we supplement with more detailed benchmark results for LRMs involved in our experiments. The fine-grained instruction-following performance across different source of math problems are presented in Table~\ref{tab:main_per_constraint}, while the hard accuracy (HAcc) and soft accuracy (SAcc) for different number of constraints are shown in Table~\ref{tab:main_per_source_hard} and Table~\ref{tab:main_per_source_soft}, respectively.

\begin{table}[h!]
    \centering
    \small
    \setlength{\tabcolsep}{6pt}
    \renewcommand{\arraystretch}{0.72}
    \caption{Experimental results of LRMs on \benchmarkname. We report hard accuracy (HAcc) and soft accuracy (SAcc) for instruction-following. $\dagger$ indicates models trained by supervised fine-tuning only (no reasoning-oriented RL).} %\textbf{Bold} and \underline{underlined} values denote the \textit{top}-2 and \textit{bottom}-2 entries in each column, respectively.}
    \renewcommand{\arraystretch}{1.05}
    \resizebox{0.7\textwidth}{!}{
% \usepackage{multirow}

\begin{tabular}{    l
    >{\columncolor{colA}}c
    >{\columncolor{colB}}c
    >{\columncolor{colB}}c
    >{\columncolor{colC}}c
    >{\columncolor{colC}}c}
\toprule
\multicolumn{1}{l}{\multirow{2}{*}{Model}}                      & single & \multicolumn{2}{>{\columncolor{colB}}c}{double} & \multicolumn{2}{>{\columncolor{colC}}c}{triple} \\
% \cmidrule(lr){3-4}\cmidrule(lr){5-6}
 & Acc    & HAcc         & SAcc        & HAcc         & SAcc        \\
\midrule
\multicolumn{6}{c}{Models with no more than 4B parameters}\\
\midrule
Qwen3-4B                                 & 53.57  & 38.57        & 57.86       & 40.00           & 72.86    \\
Qwen3-1.7B                               & 42.14  & 22.86        & 46.43       & 25.71        & 62.14       \\
Qwen3-0.6B                               & 48.57  & 22.86        & 48.93       & 12.14        & 53.81       \\
L1-Qwen-1.5B-Exact                        & 33.57  & 18.57        & 43.57       & 7.14         & 41.66       \\
L1-Qwen-1.5B-Max                          & 37.14  & 16.43        & 43.93       & 5.71         & 37.14       \\
DeepSeek-R1-Distill-Qwen-1.5B$\dagger$             & 33.57  & 14.29        & 38.21       & 3.57         & 38.09       \\
DeepScaler-1.5B-Preview                   & 30.71  & 10.00        & 35.00       & 2.86         & 37.85       \\
Qwen2.5-1.5B-SimpleRL-Zoo                 & 21.43  & 2.86         & 21.07       & 2.86         & 30.48       \\
Qwen2.5-Math-1.5B-Instruct               & 19.29  & 2.14         & 19.64       & 1.43         & 25.24       \\
\midrule
\multicolumn{6}{c}{Models with approximately 7B–14B parameters}\\
\midrule
Qwen3-14B                                & 63.57  & 40.71        & 60.71       & 47.86        & 76.90       \\
DeepSeek-R1-Distill-Qwen-14B$\dagger$              & 57.14  & 35.71        & 62.86       & 25.00         & 61.66       \\
Qwen3-8B                                 & 51.43  & 31.43        & 54.64       & 30.71        & 65.95       \\
DeepSeek-R1-Distill-Qwen-7B$\dagger$              & 39.29  & 27.14        & 50.36       & 12.86        & 45.23       \\
DeepSeek-R1-Distill-Llama-8B$\dagger$              & 34.29  & 22.14        & 47.14       & 10.00         & 50.7        \\
Open-Reasoner-Zero-7B                     & 25.71  & 13.57        & 39.64       & 1.43         & 31.42       \\
Qwen2.5-Math-7B-Instruct                 & 22.86  & 2.86         & 24.64       & 1.43         & 29.29       \\
\midrule
\multicolumn{6}{c}{Models with 32B or more parameters}\\
\midrule
Qwen3-32B                                & 61.43  & 35.00         & 57.50       & 35.00           & 69.52       \\
DeepSeek-R1-Distill-Qwen-32B$\dagger$              & 57.14  & 37.14        & 60.36       & 33.57        & 65.23       \\
DeepSeek-R1-Distill-Llama-70B$\dagger$             & 54.29  & 39.29        & 61.07       & 30.71        & 67.85       \\
QwQ-32B            & 55.71  & 35.71        & 58.57       & 29.29        & 65.69       \\
OlympicCoder-32B$\dagger$                          & 55.71  & 31.43        & 60.36       & 20.71        & 57.85       \\
s1-32B$\dagger$                                    & 37.14  & 13.57        & 38.93       & 12.14        & 49.27       \\
Open-Reasoner-Zero-32B                    & 30.71  & 13.57        & 41.79       & 2.14         & 34.05       \\
%qwen-2.5-7b-simplrl-zoo                   & 27.86  & 2.86         & 26.07       & 5.71         & 35.48      \\
\bottomrule
\end{tabular}
    }
    \label{tab:main_per_constraint}
\end{table}

\begin{table}[h]
    \centering
    %\small
    %\setlength{\tabcolsep}{6pt}
    %\renewcommand{\arraystretch}{0.72}
    \caption{Experimental results of LRMs on \benchmarkname. We report hard accuracy (HAcc) for instruction-following on five subsets of our \benchmarkname.
    %The last column shows the relative change in correctness when constraints are included. 
    %Models are sorted in descending order of overall instruction-following performance.
    $\dagger$ indicates models trained by supervised fine-tuning only (no reasoning-oriented RL).}% \textbf{Bold} and \underline{underlined} values denote the \textit{top}-2 and \textit{bottom}-2 entries in each column, respectively.}
    \renewcommand{\arraystretch}{1.05}
    \resizebox{0.75\textwidth}{!}{
% \usepackage{multirow}

\begin{tabular}{lccccc}
\toprule
\multicolumn{1}{l}{\textbf{Model}} & \textbf{GSM8K} & \textbf{MATH500} & \textbf{Minerva} & \textbf{Olympiad} & \textbf{AIME}  \\
\midrule
\multicolumn{6}{c}{Models with no more than 4B parameters} \\
\midrule
Qwen3-4B                                 & 66.67                        & 40.00                         & 53.33                          & 31.11                           & 21.67                        \\
Qwen3-1.7B                               & 44.44                        & 25.56                          & 41.11                          & 24.44                           & 8.33                         \\
Qwen3-0.6B                               & 36.67                        & 25.56                          & 34.44                          & 24.44                           & 13.33                        \\
L1-Qwen-1.5B-Exact                        & 27.78                        & 15.56                          & 21.11                          & 17.78                           & 15.00                          \\
L1-Qwen-1.5B-Max                          & 24.44                        & 18.89                          & 22.22                          & 16.67                           & 15.00                        \\
DeepSeek-R1-Distill-Qwen-1.5B$\dagger$             & 32.22                        & 12.22                          & 15.56                          & 12.22                           & 11.67                        \\
DeepScaler-1.5B-Preview                   & 26.67                        & 10.00                          & 15.56                          & 7.78                            & 11.67                        \\
Qwen2.5-1.5B-SimplRL-Zoo                 & 11.11                        & 10.00                           & 11.11                          & 4.44                            & 8.33                         \\
Qwen2.5-Math-1.5B-Instruct               & 8.89                         & 5.56                           & 8.89                           & 6.67                            & 8.33                         \\
\midrule
\multicolumn{6}{c}{Models with approximately 7B–14B parameters} \\
\midrule
Qwen3-14B                                & 71.11                        & 53.33                          & 63.33                          & 35.56                           & 20.00                           \\
DeepSeek-R1-Distill-Qwen-14B$\dagger$              & 55.56                        & 35.56                          & 44.44                          & 31.11                           & 25.00                         \\
Qwen3-8B                                 & 56.67                        & 37.78                          & 44.44                          & 24.44                           & 20.00                           \\
DeepSeek-R1-Distill-Qwen-7B$\dagger$               & 46.67                        & 22.22                          & 31.11                          & 14.44                           & 13.33                        \\
DeepSeek-R1-Distill-Llama-8B$\dagger$              & 41.11                        & 18.89                          & 20.00                             & 13.33                       & 15.00                           \\
Open-Reasoner-Zero-7B                     & 13.33                        & 14.44                          & 11.11                          & 13.33                           & 16.67                        \\
Qwen2.5-Math-7B-Instruct                 & 12.22                        & 5.56                           & 10.00                             & 8.89                            & 8.33                         \\
\midrule
\multicolumn{6}{c}{Models with 32B or more parameters} \\
\midrule
Qwen3-32B                                & 73.33                        & 40.00                         & 52.22                          & 26.67                           & 18.33                        \\
DeepSeek-R1-Distill-Qwen-32B$\dagger$              & 57.78                        & 38.89                          & 52.22                          & 32.22                           & 26.67                        \\
DeepSeek-R1-Distill-Llama-70B$\dagger$             & 55.56                        & 42.22                          & 53.33                          & 28.89                           & 20.00                           \\
QwQ-32B            &  60.00  & 38.89   & 45.56   & 32.22    &  16.67 \\
OlympicCoder-32B$\dagger$                          & 36.67                        & 36.67                          & 37.78                          & 31.11                           & 38.33                        \\
s1-32B$\dagger$                                    & 33.33                        & 20.00                             & 22.22                          & 13.33                           & 13.33                        \\
Open-Reasoner-Zero-32B                    & 15.56                        & 14.44                          & 15.56                          & 14.44                           & 18.33                        \\
\bottomrule
\end{tabular}
    }
    \label{tab:main_per_source_hard}
\end{table}

\begin{table}[h!]
    \centering
    %\small
    %\setlength{\tabcolsep}{6pt}
    %\renewcommand{\arraystretch}{0.72}
    \caption{Experimental results of LRMs on \benchmarkname. We report soft accuracy (SAcc) for instruction-following on five subsets of our \benchmarkname.
    %The last column shows the relative change in correctness when constraints are included. 
    %Models are sorted in descending order of overall instruction-following performance.
    $\dagger$ indicates models trained by supervised fine-tuning only (no reasoning-oriented RL).} %\textbf{Bold} and \underline{underlined} values denote the \textit{top}-2 and \textit{bottom}-2 entries in each column, respectively.}
    \renewcommand{\arraystretch}{1.05}
    \resizebox{0.75\textwidth}{!}{
% \usepackage{multirow}

\begin{tabular}{lccccc}
\toprule
\textbf{Model} & \textbf{GSM8K} & \textbf{MATH500} & \textbf{Minerva} & \textbf{Olympiad} & \textbf{AIME} \\
\midrule
\multicolumn{6}{c}{Models with no more than 4B parameters} \\
\midrule
Qwen3-4B                                 & 80.19                                            & 57.41                                              & 70.37                                              & 50.19                                               & 42.78                                           \\
Qwen3-1.7B                               & 65.74                                            & 44.81                                              & 61.85                                              & 45.19                                               & 25.28                                           \\
Qwen3-0.6B                               & 61.3                                             & 47.04                                              & 59.07                                              & 45.37                                               & 33.89                                           \\
L1-Qwen-1.5B-Exact                        & 50.56                                            & 37.59                                              & 39.62                                              & 33.7                                                & 35                                              \\
L1-Qwen-1.5B-Max                          & 45.37                                            & 40.56                                              & 42.78                                              & 34.44                                               & 31.1                                            \\
DeepSeek-R1-Distill-Qwen-1.5B$\dagger$             & 54.26                                            & 32.59                                              & 37.03                                              & 28.7                                                & 27.5                                            \\
DeepScaler-1.5B-Preview                   & 49.44                                            & 32.96                                              & 33.89                                              & 25.56                                               & 28.88                                           \\
Qwen2.5-1.5B-SimplRL-Zoo                 & 25.93                                            & 25                                                 & 27.96                                              & 18.7                                                & 23.89                                           \\
Qwen2.5-Math-1.5B-Instruct               & 22.41                                            & 19.07                                              & 23.33                                              & 20.37                                               & 21.94                                           \\
\midrule
\multicolumn{6}{c}{Models with approximately 7B–14B parameters} \\
\midrule
Qwen3-14B                                & 83.33                                            & 68.52                                              & 77.96                                              & 55.56                                               & 41.39                                           \\
DeepSeek-R1-Distill-Qwen-14B$\dagger$              & 76.84                                            & 58.14                                              & 62.22                                              & 55.56                                               & 44.72                                           \\
Qwen3-8B                                 & 74.44                                            & 55.74                                              & 64.07                                              & 45                                                  & 42.5                                            \\
DeepSeek-R1-Distill-Qwen-7B$\dagger$               & 67.96                                            & 41.67                                              & 52.59                                              & 29.44                                               & 27.22                                           \\
DeepSeek-R1-Distill-Llama-8B$\dagger$              & 62.59                                            & 42.22                                              & 43.51                                              & 35.93                                               & 31.93                                           \\
Open-Reasoner-Zero-7B                     & 32.22                                            & 32.78                                              & 31.67                                              & 29.62                                               & 36.38                                           \\
Qwen2.5-Math-7B-Instruct                 & 29.63                                            & 20.93                                              & 27.41                                              & 25.19                                               & 24.44                                           \\
\midrule
\multicolumn{6}{c}{Models with 32B or more parameters} \\
\midrule
Qwen3-32B                                & 86.11                                            & 59.26                                              & 70.74                                              & 48.89                                               & 42.22                                           \\
DeepSeek-R1-Distill-Qwen-32B$\dagger$              & 75.73                                            & 60.37                                              & 67.78                                              & 50.73                                               & 44.44                                           \\
DeepSeek-R1-Distill-Llama-70B$\dagger$             & 75.73                                            & 60.93                                              & 70.56                                              & 48.89                                               & 43.33                                           \\
QwQ-32B            &  78.14                     &  57.03                       &  66.67                       &  51.11                        &  40.50                     \\
OlympicCoder-32B$\dagger$                        & 58.89                                            & 55.92                                              & 64.26                                              & 54.26                                               & 55.83                                           \\
s1-32B$\dagger$                                    & 54.81                                            & 43.51                                              & 45.56                                              & 31.48                                               & 29.43                                           \\
Open-Reasoner-Zero-32B                    & 36.85                                            & 33.52                                              & 37.04                                              & 33.15                                               & 37.78                                           \\
\bottomrule
%qwen-2.5-7b-simplrl-zoo                   & 33.89                                            & 29.81                                              & 29.63                                              & 27.78                                               & 26.94                                          
\end{tabular}
    }
    \label{tab:main_per_source_soft}
\end{table}

\section{More Results on Math Benchmarks}
\label{sec:more_math}
In Section 5.3, we control the CoT length during RL training and report the averaged math reasoning performance among five benchmarks (AIME2024, AIME2025, AMC2023, Minerva, and Olympiad) in Table 5. For fine-grained analysis, we report more detailed results on five benchmarks in Table~\ref{tab:table/rollout_math}.

\begin{table}[h!]
    \centering
    \caption{Reasoning performance for LRMs when trained with varying maximum response length (the number in the bracket) during RL.}
    \resizebox{0.9\linewidth}{!}{
    \begin{tabular}{lcccccc}
\toprule
\textbf{Model}                                  & \textbf{AIME2024} & \textbf{AIME2025} & \textbf{AMC2023} & \textbf{Minerva} & \textbf{Olympiad} & \textbf{Average} \\
\midrule
Original     & 28.33    & 21.15    & 67.73   & 23.16   & 40.30    & 36.13   \\
\midrule
\quad+cold-RL (1k) & 14.27    & 11.67    & 58.20   & 23.53   & 36.00    & 28.73   \\
\quad+cold-RL (2k) & 24.06    & 19.58    & 70.39   & 26.10   & 41.48    & 36.32   \\
\quad+cold-RL (4k) & 28.65    & 24.17    & 75.39   & 26.47   & 45.48    & 40.03   \\
\quad+cold-RL (8k) & 30.73    & 24.06    & 73.05   & 26.84   & 44.44    & 39.82  \\
\bottomrule
\end{tabular}
    }
    \label{tab:table/rollout_math}
\end{table}

\begin{table}[h!]
    \centering
    \caption{Reasoning performance for LRMs when trained with different reasoning-oriented training strategies.}
    \resizebox{0.9\linewidth}{!}{
    
\begin{tabular}{lcccccc}
\toprule
                  & \textbf{AIME2024} & \textbf{AIME2025} & \textbf{AMC2023} & \textbf{Minerva} & \textbf{Olympiad} & \textbf{Average}  \\
\midrule
Qwen2.5-1.5B      & 0.21     & 0.00        & 2.89    & 1.47    & 1.48     & 1.21  \\
+SFT              & 0.10     & 0.10     & 10.70   & 4.04    & 6.07     & 4.20  \\
+SFT+RL           & 4.48	 & 2.08	    &28.36	  & 9.56	& 18.22     & 12.54  \\
+cold-RL          & 4.48     & 2.19     & 30.47   & 16.18   & 19.56    & 14.58 \\
\quad w/ format reward   & 2.60     & 0.31     & 26.80   & 9.56    & 16.59    & 11.17 \\
\midrule
Qwen2.5-7B        & 4.90     & 1.98     & 27.81   & 13.24   & 20.00    & 13.59 \\
+SFT              & 10.00    & 10.52    & 40.78   & 25.00   & 29.19    & 23.10 \\
+SFT+RL           & 18.65	 & 18.23	& 57.34	  & 27.94	& 41.93    & 32.82 \\
+cold-RL          & 15.21    & 8.75     & 53.98   & 29.78   & 34.22    & 28.39 \\
\quad w/ format reward   & 10.52    & 8.13     & 46.56   & 27.21   & 31.56    & 24.80 \\
\midrule
Qwen2.5-Math-1.5B & 7.92     & 4.27     & 42.89   & 14.71   & 24.74    & 18.91 \\
+SFT              & 5.94     & 3.65     & 30.08   & 13.60   & 18.67    & 14.39 \\
+SFT+RL           & 10.94	 & 9.27	    & 48.75	  & 23.16	& 31.41    & 24.71 \\
+cold-RL          & 13.30    & 7.70     & 52.00   & 20.58   & 30.81    & 24.88 \\
\quad w/ format reward   & 12.81    & 6.46     & 51.95   & 20.22   & 28.30    & 23.95 \\
\midrule
Qwen2.5-Math-7B   & 16.45    & 8.13     & 45.63   & 7.72    & 25.48    & 20.68 \\
+SFT              & 16.88    & 15.94    & 53.36   & 25.00   & 34.37    & 29.11 \\
+SFT+RL           & 30.21	 & 23.96	& 70.55	  & 31.25	& 47.26    & 40.65 \\
+cold-RL          & 27.50    & 13.60    & 59.84   & 25.36   & 36.74    & 32.61 \\
\quad w/ format reward   & 28.75    & 11.15    & 62.50   & 26.10   & 34.81    & 32.66 \\
\bottomrule
\end{tabular}
    }
    \label{tab:my_label}
\end{table}

\section{More Results on Previous Instruction-following Benchmarks.}
\label{sec:more_if}
As shown in Figure 1 and Table~\ref{tab:ifeval_three_pairs}, we evaluate the instruction-following performance of three LRMs together with the Instruction models from which they are trained, and report the results on IFEval~\cite{ifeval} and FollowBench~\cite{followbench}. From the table, we can observe that reasoning-oriented training seems to be detrimental to instruction-following. %The performance for other LRMs involved in our benchmark on IFEval are presented in Table~\ref{tab:table/ifeval_all}.  

\begin{table}[h]
    \centering
    \caption{Instruction-following performance for three Instruction models and LRMs trained on them (highlighted in gray). We report the prompt-level strict accuracy for IFEval and use GPT-4o-mini as evaluator in FollowBench.}
    \resizebox{0.5\linewidth}{!}{
    \rowcolors{2}{gray!10}{}
    
\begin{tabular}{lcc}
\toprule
                              & \textbf{IFEval} & \textbf{FollowBench} \\
\midrule
Llama-3.3-70B-Instruct        & 90.38   & 61.82        \\
DeepSeek-R1-Distill-Llama-70B & 78.74   & 49.26        \\
\midrule
Qwen2.5-32B-Instruct         & 80.96   & 60.12        \\
s1-32B & 58.04   & 51.46        \\
\midrule
Qwen2.5-Coder-32B-instruct   & 80.03   & 59.14        \\
OlympicCoder-32B              & 57.11   & 46.70      \\
\bottomrule
\end{tabular}

    }
    \label{tab:ifeval_three_pairs}
\end{table}

% \begin{table}[]
%     \centering
%     \caption{Caption}
%     \resizebox{0.6\linewidth}{!}{
%     \input{table/ifeval_all}    
%     }
%     \label{tab:table/ifeval_all}
% \end{table}

\section{List of Constraints}
\label{sec:constraint_list}
In this section we provide a detailed list of the $15$ constraints used in our benchmark in Table~\ref{tab:table/constraint_list}.

\begin{table}[t]
  \centering
  \small
  \setlength{\tabcolsep}{5pt}
  \renewcommand{\arraystretch}{1.0}
  \caption{The list of $15$ constraints used in our proposed \benchmarkname. }
  \label{tab:table/constraint_list}
  \vspace{1pt}
  \resizebox{0.9\textwidth}{!}{%
% \usepackage{multirow}
\begin{tabular}{
 >{\bfseries}l  % 1st column: left-aligned + bold text (applies \bfseries to every cell)
 >{}p{0.8\linewidth} % 3rd column: fixed width (80% of linewidth) with automatic line breaks
}
\toprule
Category                 & \textbf{Constraint}                                                                                                       \\
\midrule
length                   & $\bullet$ Answer with at least/around/most \{N\} words.                                                             \\
\midrule
\multirow[t]{4}{*}{lexical} & $\bullet$  Include keywords \{keyword1\}, \{keyword2\} in your response.                                                     \\
                         & $\bullet$ In your response, the word word should appear \{N\} times.                                                       \\
                         & $\bullet$ Do not include keywords \{forbidden words\} in the response.                                                     \\
                         & $\bullet$ Your ENTIRE response should be in \{language\}, no other language is allowed. \\
\midrule
\multirow[t]{7}{*}{format}  & $\bullet$  Your answer must contain exactly \{N\} bullet points. Use the markdown bullet points such as: * This is a point. \\
                         & $\bullet$  Highlight at least \{N\} sections in your answer with markdown, i.e. highlighted section.                         \\
                         & $\bullet$  Your response must have \{N\} sections. Mark the beginning of each section with \{section\_splitter\} X.         \\
                         & $\bullet$  Your entire response should be in English, capital letters only.                                                 \\
                         & $\bullet$  Your entire response should be in English, and in all lowercase letters. No capital letters are allowed.         \\
                         & $\bullet$  In your response, words with all capital letters should appear at least / around / at most \{N\} times.          \\
                         & $\bullet$  In your entire response, refrain from the use of any commas.                                                     \\
\midrule
\multirow[t]{3}{*}{affix}   & $\bullet$  Finish your response with this exact phrase \{end\_phrase\}. No other words should follow this phrase.           \\
                         & $\bullet$  Wrap your entire response with double quotation marks.                                                           \\
                         & $\bullet$  First, repeat the request without change, then give your answer. \\
\bottomrule
\end{tabular}
  } % end resizebox
\end{table}

% \begin{table}[]
%     \centering
%     \caption{The reasoning performance on math benchmarks for different reasoning-oriented training strategies and different backbones.}
%     \resizebox{0.85\linewidth}{!}{
%     \input{table/reasoning_training_math}
%     }
%     \label{tab:reasoning_training_math}
% \end{table}

% \section{Other Appendix Sections}
% For other appendix sections (e.g., detailed training setup, model configuration and detailed model evaluation), please see supplementary materials.

\label{sec:limitations}

% \input{checklist}

%%%%%%%%%%%%%%%%%%%%%%%%%%%%%%%%%%%%%%%%%%%%%%%%%%%%%%%%%%%%

% \appendix

% \section{Technical Appendices and Supplementary Material}
% Technical appendices with additional results, figures, graphs and proofs may be submitted with the paper submission before the full submission deadline (see above), or as a separate PDF in the ZIP file below before the supplementary material deadline. There is no page limit for the technical appendices.

%%%%%%%%%%%%%%%%%%%%%%%%%%%%%%%%%%%%%%%%%%%%%%%%%%%%%%%%%%%%

\end{document}